\documentclass[10pt,twocolumn,letterpaper]{article}
\usepackage{cvpr}
\usepackage{times}
\usepackage{epsfig}
\usepackage{graphicx}
\usepackage{amsmath}
\usepackage{amssymb}
\usepackage{pifont}
\usepackage{multirow}
\newcommand{\cmark}{\ding{51}}%
\newcommand{\xmark}{\ding{55}}%

\usepackage[pagebackref=true,breaklinks=true,letterpaper=true,colorlinks,bookmarks=false]{hyperref}

\cvprfinalcopy % *** Uncomment this line for the final submission

\graphicspath{{figures/}}
\def\myparagraph#1{\medskip\noindent{\bf #1}~~}

\ifcvprfinal\fi
\begin{document}

%%%%%%%%% TITLE
\title{
Restoring Images with Unknown Degradation Factors\\
by Recurrent Use of a Multi-branch Network}
% Task-Blind Image Restoration via Multi-task Trained Network}

\author{Xing Liu\textsuperscript{\textdagger} ~~~~~Masanori Suganuma\textsuperscript{\textdagger\textdaggerdbl} ~~~~~Xiyang Luo\textsuperscript{\textdagger} ~~~~~Takayuki Okatani\textsuperscript{\textdagger\textdaggerdbl} \\
\textsuperscript{\textdagger}Graduate School of Information Sciences, Tohoku University ~~~~~~~~~~\textsuperscript{\textdaggerdbl}RIKEN Center for AIP \\
{\tt\small \{ryu,suganuma,luo,okatani\}@vision.is.tohoku.ac.jp}
}

\maketitle
%\thispagestyle{empty}
%%%%%%%%% ABSTRACT
\begin{abstract} 
The employment of convolutional neural networks has achieved unprecedented  performance in the task of image restoration for a variety of degradation factors. However, high-performance networks have been specifically designed for a single degradation factor. In this paper, we tackle a harder problem, restoring a clean image from its degraded version with an unknown degradation factor, subject to the condition that it is one of the known factors. Toward this end, we design a network having multiple pairs of input and output branches and use it in a recurrent fashion such that a different branch pair is used at each of the recurrent paths. We reinforce the shared part of the network with improved components so that it can handle different degradation factors. We also propose a two-step training method for the network, which consists of multi-task learning and fine-tuning. 
% This design enables to adopt the fomulation of image restoration that is known to be effective (i.e., predicting the difference between the degraded input and the ideal clean image) as well as to significanly reduce memory consumption as compared with applying dedicated networks for different factors in turn. 
The experimental results show that the proposed network yields at least comparable or sometimes even better performance on four degradation factors as compared with the best dedicated network for each of the four. We also test it on a further harder task where the input image contains multiple degradation factors that are mixed with unknown mixture ratios, showing that it achieves better performance than the previous state-of-the-art method designed for the task. 
% \color{blue}
% Convolutional neural networks have recently been successfully applied to the problems of restoring clean images from their degraded versions. Most studies have designed and trained a dedicated network for each of many image restoration tasks, such as motion blur removal, rain-streak removal, haze removal, etc. In this paper, we show that a single network having multiple input and output branches can solve multiple image restoration tasks, without knowing the distortion type in advance for each image. 
% This is made possible by i) improving the attention mechanism and an internal structure of the basic blocks used in the dual residual networks, which was recently proposed and shown to work well for a number of image restoration tasks; ii) a task-recurrent training scheme that makes the network a universal solution for the tasks. Experimental results show that the proposed approach generally outperforms the previous approaches. To the authors' knowledge, this is the first report of accurate \underline{blind} image restoration on \underline{diverse} distortion types. 
\end{abstract}

%%%%%%%%% BODY TEXT

\section{Introduction}
\label{sec:intro}
The problem of image restoration, i.e., restoring an original, clean image from its degraded version(s), has been studied for a long time in computer vision and image processing. As with other problems of computer vision, deep learning has been applied to this problem, leading to significant improvement of performance. There are many factors causing image degradation, for each of which there are a large number of studies in the past, such as motion/defocus blur \cite{classic_deblur1,classic_deblur2,xu-blur,deblur3,classic_defocus_blur}, several types of noises (e.g., Gaussian, real-world noise, etc.) \cite{bm3d, pclr, pgpd, twsc}, JPEG compression noise \cite{classic_jpeg0,classic_jpeg1,classic_jpeg2,classic_jpeg3}, rain streak \cite{noncnn_derain1,noncnn_derain2,noncnn_derain3}, raindrop \cite{raindrop-relate1,raindrop-relate2,raindrop-relate3}, haze \cite{kaiming_dehaze,dehaze_relate2,berman_dehaze} etc.

Previous studies have treated each of these degradation factors individually and developed ``dedicated'' methods for each factor. It is also the case with recent studies utilizing deep learning\cite{DeBlurGAN,raindrop18,god_suganuma,rescan,jpeg_related2,dehaze_zhanghe}; 
%there are different networks for different degradation types. 
%a network is designed and trained for each degradation factor.
This means that it needs to be known in advance which factor of degradation needs to be removed from the image. Although this will be fine for photo-editing software, where users specify it, this is insufficient for real-world applications such as self-driving/driver assistance and surveillance.
%, as it cannot be known in advance in what factor(s) the input image is degraded. 
\begin{figure}[bt]
\centering
\includegraphics[width=\columnwidth]{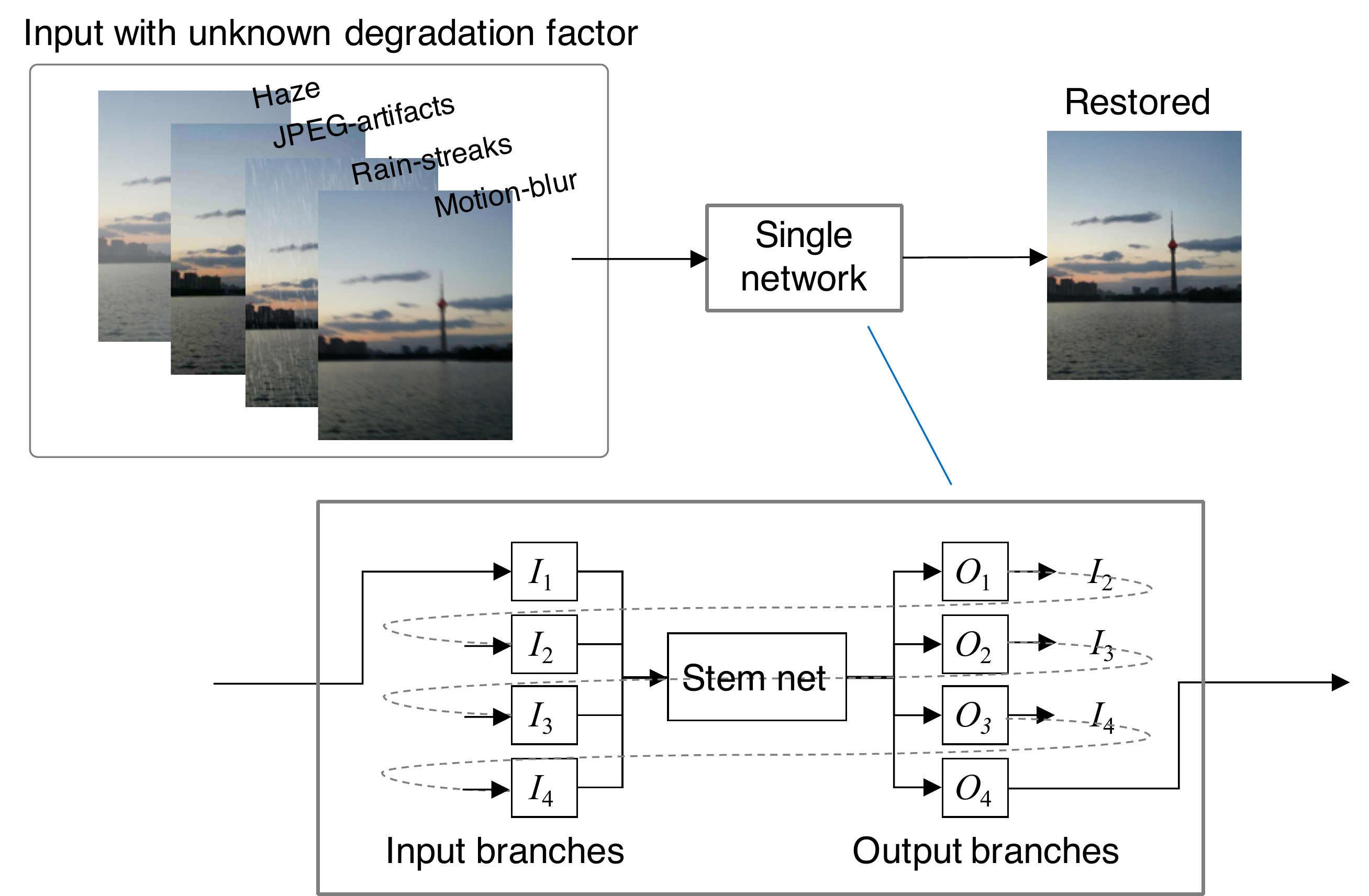}

\medskip
\caption{Illustration of the goal and our approach. 
To restore a clean image from an input with unknown degradation factor, we use a network with multiple input and output branches. Receiving an input image from $I_1$, it yields an improved image in terms of a single degradation factor from $O_1$. It is then fed to $I_2$ and improved in terms of another factor, outputting the improved image from $O_2$. This is repeated until $O_4$, yielding the final output. Each box inside the bottom large box represents a sub-network. }
% Left: Approach employed in recent studies, i.e.,  designing/training a different network for each image restoration task dealing with a single degradation factor. Right: Our approach; a single network having a single input and multiple output branches is trained on multiple image restoration tasks.}
\label{fig:face}
%\vspace{-0.3cm}
\end{figure}

Thus, it is desirable to enable to deal with input images with unknown degradation factor(s), as shown in the upper row of Fig.~\ref{fig:face}. It should be noted that a few studies tackled this problem \cite{toolchain,suganuma_cvpr19_arxiv}. However, their performances are no so high; they are significantly lower than those of dedicated networks used in the ideal case when they are applied to images having the assumed degradation factor.

\begin{figure}[bt]
\centering
\includegraphics[width=0.85\columnwidth]{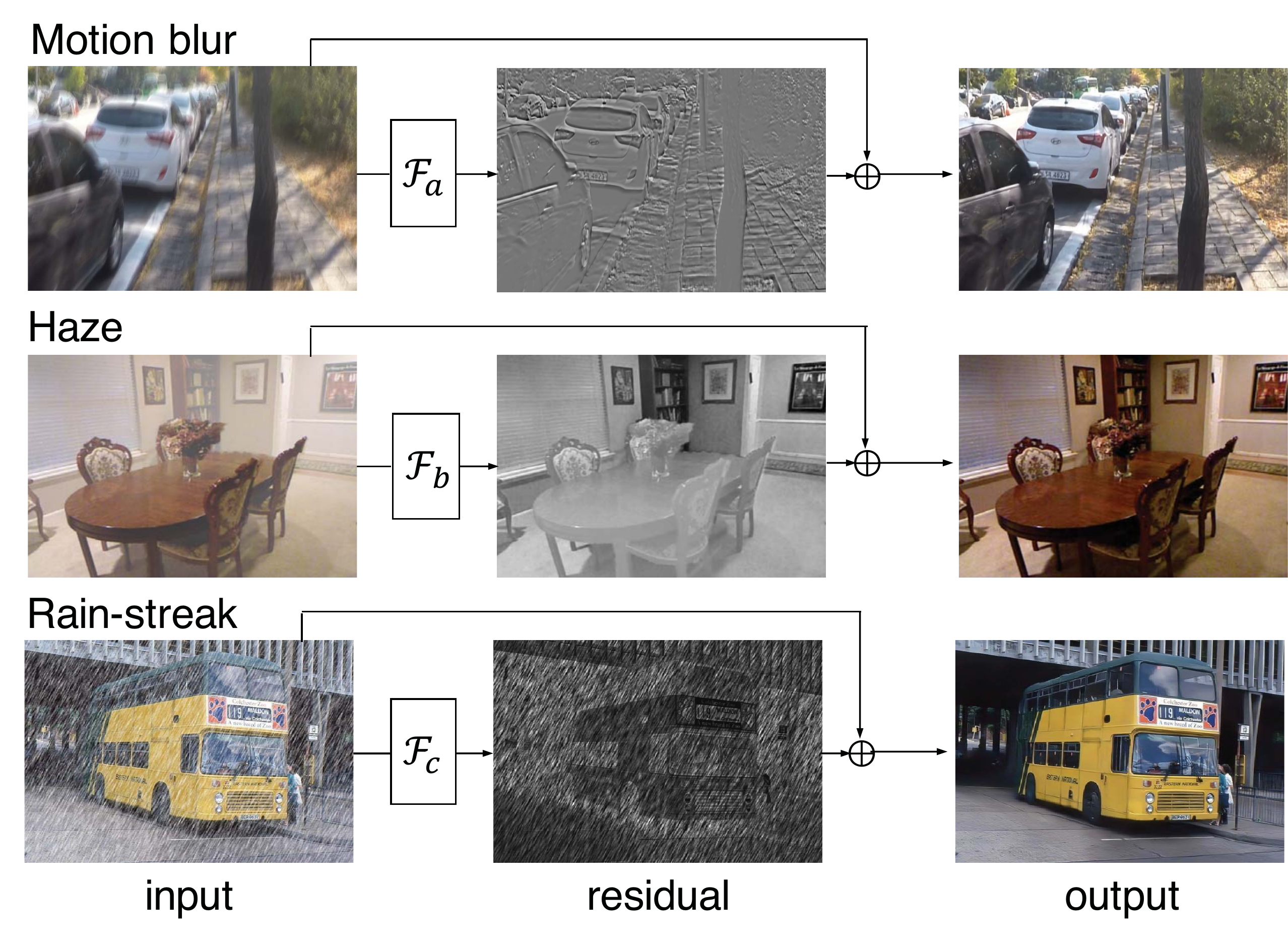}
\caption{The approach that is currently very successful for image restoration tasks. CNNs $\mathcal{F}'s$ are used to predict the difference (or residue) from an input to its ideally restored version (i.e., a clean image). A single network is designed and trained for each task.}
% $\mathcal{F}_{a}$, $\mathcal{F}_{b}$ and $\mathcal{F}_{c}$ denotes a CNN trained for motion blur removal, haze removal and rain-streak removal. The images in the second column are normalized for better visibility.}
\label{fig:outermost}
%\vspace{-0.3cm}
\end{figure}

How can we restore images with unknown degradation factors to a higher level? We can think of two approaches. One is to build a universal network that can deal with any degradation factors\footnote{In this paper, we consider only degradation factors for which training data are available. Completely novel factors are beyond the scope.}; it is a network equipped with a single input receiving a degraded image and a single output yielding a restored image. However, such a design has  difficulty. A very successful approach to image restoration so far is 
%; it is difficult to have a single output layer deal with different degradation factors.
to have a network predict only the difference from the input image to the desired high-quality image \cite{DuRN,tianda_rain,rescan}, as shown in Fig.~\ref{fig:outermost}. As seen in the figure, the residual images  tend to have considerably different statistical properties for different factors, making it difficult to have a single output deal with them. 

The other approach is to cascade multiple networks, each of which is a dedicated network for a single factor. It should perform well if each component network works well on the target factor and also does no harm on the other factors as well as restored clean images. 
%However, we found through our experiments that this approach does not work so well. It seems that the above assumption does not hold true for existing networks for image restoration tasks. 
This approach, however, will suffer from a high memory consumption; the total number of parameters in the cascade increases proportionally with the number of factors we consider, leading to excessive memory use in the case of many factors. 

In this paper, we consider yet another (the third) approach, which may be considered as an intermediate solution between the above two. We consider a single network with multiple input and output branches, as shown in Fig.~\ref{fig:face}. Each output branch forms a pair with one of the input branches, which is used for removing a single degradation factor. Then, we recurrently use this network, that is, we feed the output from an output branch back to one of the input branches and repeat this procedure for the number of factors with a different input-output pair at each time. (This may be viewed as a cascading repetition of the same network with each cascading point connected with a different input-output pair.) 
To train the network, we consider a two-step method; in the first step, we train it in a multi-task learning (MTL) framework, and in the second step, we fine-tune it so that it can be used in a recurrent manner. 
%To be specific, we employ a task scheduling method to do this; the network is trained on one of the target restoration tasks at a time and this is iterated for many times while switching tasks. 

This approach resolves the difficulties above with the two approaches. As the network has multiple input and output branches, we can employ the standard formulation of predicting a residual image for each factor. Our training scheme enables us to equip the network with the property that it does not affect the factors other than the target one including  clean images. 

Although our network has multiple input and output branches, its stem, i.e., the shared main body, is dominant in terms of the number of parameters. To provide it with a sufficient representational power to deal with multiple degradation factors, we propose a new design of a network component built upon the {\em dual residual networks}, which was recently proposed by Liu \etal \cite{DuRN}. Although the authors have shown that the base network architecture is effective for various degradation factors, its internal components need to be changed for different degradation factors. We present two improvements to its basic block. One is an improved attention mechanism that utilizes amplitude of spatial derivatives of activation (i.e., $\vert I^{(c)}_x\vert$ and $\vert I^{(c)}_y\vert$) 
%in addition to activation itself (i.e., $I^{(c)}$) 
to compute attention weights. The other is a new internal design, in which the two operations employed in DuRB-U and -US \cite{DuRN} are fused; we will refer to it as DuRB-M.
%DuRB-S and -US. 
We show through experiments that these two improvements do contribute to achieving the goal of this study. 

% {\color{green}
% They employed the squeeze-and-excitation (SE) mechanism in some of their proposed component blocks, named DuRB-S and DuRB-US. Since it was originally proposed for object classification task, the SE mechanism has been successfully applied to various tasks such as super-resolution (SR) \cite{rcan}, single-view depth estimation \cite{ko-san_depth}, etc. Its concept is to use global average pooling of activations in each individual channel of a layer to generate channel-wise attention weights on its layer activations. We extend it to additionally use amplitude of spatial derivatives of activation (i.e., $\vert I^{(c)}_x\vert$ and $\vert I^{(c)}_y\vert$) 
% %in addition to activation itself (i.e., $I^{(c)}$) 
% to compute attention weights. The other extension is a new design of basic block, in which the two operations employed in DuRB-U and -US \cite{DuRN} are fused; we will refer to it as DuRB-M.
% %DuRB-S and -US. 
% We show through experiments that these two extensions improve the performance on multiple restoration tasks by a large margin.
% }

\section{Related Work}

\paragraph{Image Restoration} 
Image restoration has been studied for a long time. Most of the early studies incorporate models of degraded images along with priors of clean natural images, based on which they formulate and solve an optimization problem. Examples are the studies on motion blur removal \cite{Fergus,xu-blur,deblur3,deblur4} and those on haze removal \cite{kaiming_dehaze,dehaze_relate1,dehaze_relate2}. 
Recently, CNN-based methods have achieved good performance for all sorts of image restoration tasks \cite{sun-blur,gong-blur,nah,dehaze_zhanghe,Yang_2018_ECCV,Ren_dehaze,derain_zhanghe,rescan,derain_sota_acm,nonlocal_iclr,D3_jpeg,ARCNN_jpeg,multi-baseline1}. For motion blur removal, Nah \etal \cite{nah} proposed a network architecture having modified residual blocks and trained it on a large scale dataset (GoPro Data). Kupyn \etal \cite{DeBlurGAN} proposed a GAN \cite{GAN} -based method, updating the former state-of-the-art.
For haze removal, Zhang \etal \cite{dehaze_zhanghe} proposed a GAN-based CNN that jointly estimates multiple unknowns comprising a haze model. Ren \etal \cite{GFN} proposed a method of weighting several enhanced versions of an input image with the weights predicted by a CNN. 

For rain-streak removal, Li \etal proposed a RNN-based method \cite{rescan} and Li \etal proposed non-locally enhanced dense block (NEDB) \cite{derain_sota_acm}.
For JPEG-compression noise removal, Galteri \etal \cite{jpeg_related1} proposed a GAN-based method and
Zhang \etal \cite{nonlocal_iclr} proposed a non-local attention mechanism.
Finally, Liu \etal recently proposed a network architecture having dual residual connections, updating state-of-the-art performance on most of the above tasks. 

\myparagraph{Single Net for Multiple Degradation Factors}
A more challenging problem is to restore a clean image from an input image with an unknown combination of multiple degradation factors, which were studied in \cite{toolchain,suganuma_cvpr19_arxiv}. However, there seems to be a large room for improvement, because the performances of their methods are inferior with large margin to the state-of-the-art  method designed for each single factor in the removal performance of that factor. There are studies that train a network on two or more degradation factors that are similar (e.g., deblurring and super-resolution \cite{blur-SR1,deblur_and_sr}) or closely related with each other (e.g., denoising and deblur/decompression \cite{denoise-decompre,nah} and rain-streak and haze \cite{rain-and-haze}) to improve performance. These methods are specifically designed for a particular set of degradation factors; thus it will be hard to apply them to the removal of a diverse range of factors. 
There are also studies that train different networks with an identical design on different degradation factors e.g., noise, rain-streak, and super-resolution \cite{xUnit}; or noise, mosaic, JPEG-compression noise, and super-resolution \cite{nonlocal_iclr}.
Our study differs from these in the motivation explained in Sec.~\ref{sec:intro} and also in that we train the same network on multiple  factors. 
% Our study differs from these studies in motivation as described in Sec.~\ref{sec:intro}, resulting in differences in the number (up to four) and diversity of degradation factors that are simultaneously considered. 

\myparagraph{Use of Recurrent Paths for Image Restoration}
A number of studies on image-to-image translation tasks have employed the idea of using the entire or a part of a network in a recurrent fashion for several purposes, such as improved efficiency and accuracy, e.g., image compression \cite{RNN-rela1}, saliency detection \cite{RNN-rela2}, segmentation \cite{RNN-rela4}, and depth estimation \cite{RNN-rela3}. 
It has been adopted for image restoration tasks on single images \cite{dualstate,DynamicBlur,tao-blur,raindrop18} as well as videos \cite{RNN-vid2018-1,RNN-vid2018-2,RNN-vid2019-3}. 
%We summarize here only the studies on single image restoration tasks. 
%It has been witnessed that recurrently using the entire or a part of a network improves its efficiency for various image based problems, such as image compression \cite{RNN-rela1}, saliency detection \cite{RNN-rela2}, segmentation \cite{RNN-rela4} and depth estimation \cite{RNN-rela3}. 
%In image restoration tasks, this strategy is widely used on both single images \cite{dualstate,DynamicBlur,tao-blur,raindrop18} and videos \cite{RNN-vid2018-1,RNN-vid2018-2,RNN-vid2019-3}. 
%We focus on single image restoration tasks. 
For instance, it has been shown \cite{tao-blur,nest_blur,ECP_blur}  that recurrently scaling-up input images for training networks is effective for dynamic scene deblurring. 
% In this strategy, the coarse scale version (down-scaled by a factor of 4 from the original size) of an training image is firstly fed into a network to have the deblurred result. 
% Then, the result is scaled up by a factor of 2 to the middle scale, and is re-fed into the network with the middle scale version of the training image. This procedure repeats again for the fine scale version to have the final deblurring result. 
Gao \etal \cite{nest_blur} point out that blurring effects vary in magnitude depending on the scale of input images. While sharing all layers to deal with multiple scales often ends up with learning only single-scale features,  
%causes the features learnt for one scale dominates the parameters. 
%On the other hand, isolating the most or all of layers for different
assigning different layers for different scales \cite{nah,ECP_blur} is inefficient.
Thus, they propose a selective sharing approach to share all layers for multiple scales of input images except the convolutional layers at the beginning and end, and those employed for down/up-scaling. For the rain-streak removal task, it has been known to be an effective strategy to 
progressively restore a clean image from the input.
%has been known as an effective strategy for this task. 
This strategy is implemented by repeatedly sending the output of a network back to its input and performing forward computation; for this purpose, multiple gating units (\eg LSTM) is employed \cite{rescan} or one Conv-LSTM is used at the beginning of the network \cite{tianda_rain}. 
% \cite{tianda_rain,rescan}. 
% Li \etal \cite{rescan}'s implementation uses multiple gating units (\eg LSTM) in their network, while Ren \etal \cite{tianda_rain}'s implementation sets one Conv-LSTM at the beginning of the network. 

\myparagraph{Multi-task Learning} It has been well known that multi-task learning (MTL) \cite{mutli-task0} is effective for deep networks applied to many computer vision tasks;  \cite{mutli-task0,multi-task2,multi-task3,multi-task4,mask-rcnn,edsr} to name a few. To enable MTL to work, there should arguably be some relation among the tasks jointly learned; in other words, there should be overlaps among the representations to be learned inside networks for those tasks. 
% Combinations of tasks in the successful MTL examples include scene recognition and object recognition \cite{mask-rcnn}; depth estimation and scene parsing \cite{multi-task1}; 
% facial expression recognition and landmark detection \cite{face_and_mental};
% vision and language \cite{kien_cvpr19} \etc.
It remains unknown if MTL is effective for diverse image restoration tasks; 
%. Although there should be some similarity among them, 
different degradation factors could be orthogonal with each other. In fact, the aforementioned studies on the restoration from combined degradation factors attempt to deal with different  factors by adaptively selecting different networks \cite{toolchain} or different operations \cite{suganuma_cvpr19_arxiv} depending on the factors existing in the input image. 

% \paragraph{Squeeze-and-Excitation Mechanism for Attention} 
% Attention mechanisms have been developed and employed to solve various computer vision problems \cite{attention_related1,attention_related2,attention_related3,senet,suganuma_cvpr19_arxiv} . Hu \etal proposed a squeeze-and-excitation (SE) block, which produces and applies channel-wise weights on an input tensor \cite{senet}. This block has been successfully applied to various tasks such as classification \cite{senet}, super resolution \cite{rcan} and single-view depth estimation \cite{ko-san_depth}. A number of later studies \cite{gala,CompetSEResNet,cbam,senet2,Sec_order_pool} aim at improving the SE-block. Woo \etal propose to use channel-wise and spatial attention weights \cite{cbam}.  
% Hu \etal study how to efficiently combine a SE-block and a ResNet module \cite{CompetSEResNet}. Gao \etal  propose to use correlation of activations of each pair of channels
% %instead of (the global average pooling of) activation of each channel 
% to generate attention \cite {Sec_order_pool}. Hu \etal improve SE-block by replacing global average pooling with a pooling operation 
% with trainable parameters \cite{senet2}.

{
}

\section{Design of the Proposed Network} %for Multi-task Restoration}

As shown in Fig.~\ref{fig:face}, the proposed network consists of multiple input branches, a stem network, and multiple output branches. We first describe the design of the stem network, and then present the design of output branches and finally input branches. 

\subsection{Design of the Stem Network}

The stem network, which occupies the dominant portion in the entire network in terms of the number of parameters, is built upon the dual residual network of Liu et al. \cite{DuRN}. Its main part consists of a number of stacks of the basic blocks named DuRB-{$X$}'s. In our preliminary experiments, we tested all the variants of the basic blocks presented in their paper to build the stem network, and found that some of them work fairly well but never match the best network currently reported for each task. Aiming at achieving better performance, we propose to make two improvements in the basic block. The goal is to build a single architecture that can deal with many degradation factors. We will call the improved block {\em DuRB-M}. We build the stem network by DuRB-M for several times; we use five stacks in our experiments.

\subsubsection{Improved Attention Mechanism}
%{\color{red} [Intuition is written in line: 990-1003]}}
\label{sec:IAM}
\begin{figure}[bt]
\centering
\includegraphics[width=\columnwidth]{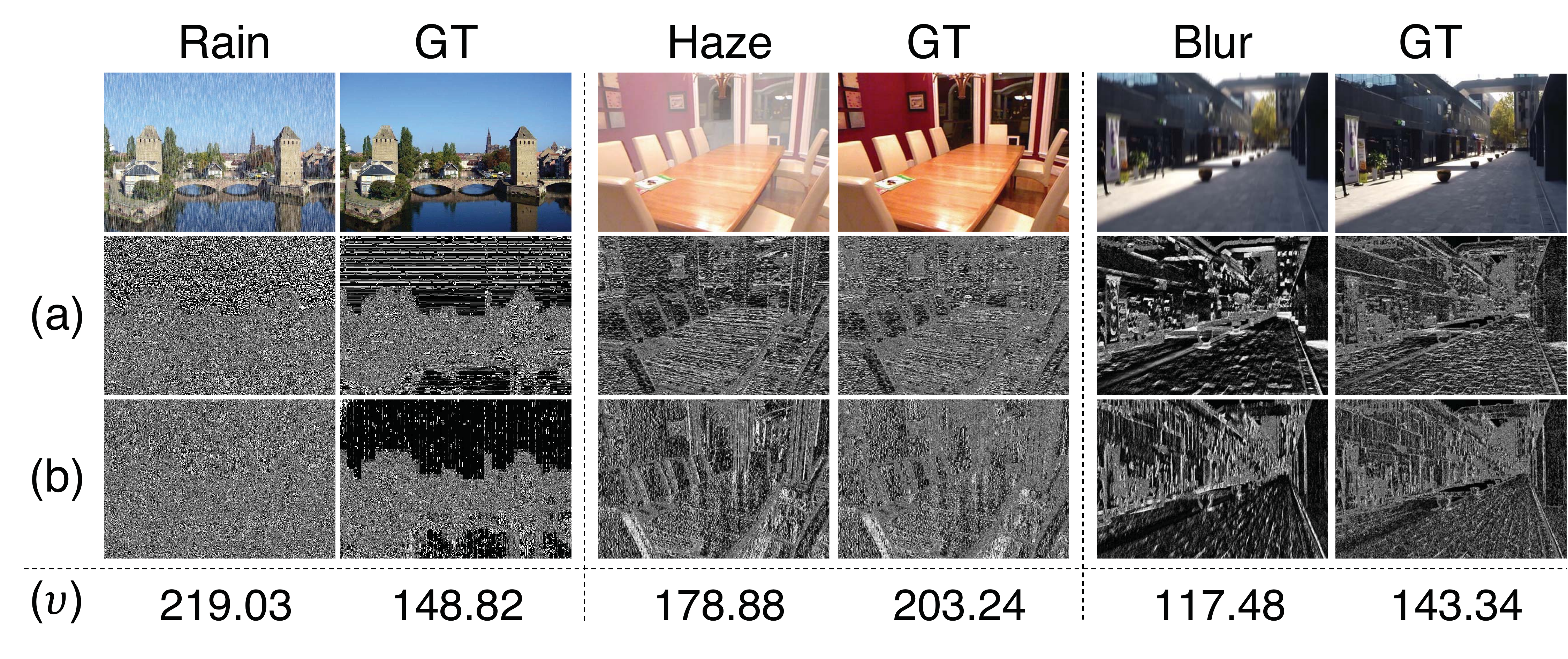}
\caption{Absolute spatial derivatives of images in (a) vertical and (b) horizontal directions. (The values in the three color channels are summed together.) The values in the bottom are computed with the application of \eqref{equ:tv} on all the three color channels. }
\label{fig:tv_image_examp}
%\vspace{-0.3cm}
\end{figure}

The dual residual networks employ an attention mechanism, which is the channel-wise attention that was originally developed for object recognition in the study of squeeze-and-excitation (SE) networks \cite{senet}, and has been widely used for many other tasks. A SE-block computes and applies attention weights on the channels of the input feature map. To determine the weight on each channel, it computes the averages of activation values of channels; then, they are converted by two fully-connected layers with ReLU and sigmoid activaton functions to generate channel-wise weights. The aggregation of activation values is equivalent to global average pooling.
We enhance this attention mechanism by incorporating a different aggregation method of channel activation. 

Our idea is to use different statistics of channel activation values in addition to their averages. For this, we choose to use (absolute) spatial derivatives of channel activation values. More specifically, denoting an activation value at spatial position $(i,j)$ of channel $c$ by $y_{c,i,j}$, we calculate  
\begin{equation} \small
\label{equ:tv}
  v_c = \frac{1}{N} \sum_{i,j} \lvert y_{c,i+1,j} - y_{c,i,j} \rvert^{\beta} + \frac{1}{M} \sum_{i,j}\lvert y_{c,i,j+1} - y_{c,i,j} \rvert^{\beta},
\end{equation}
where $N$ and $M$ denote the number of values in the corresponding spatial derivative maps; $\beta$(=3 for our experiments) is a scalar to enhance derivative values.
This is also known as the total variation \cite{first_tv}.
%, which has been used as a regularization term for various image processing tasks; a notable example is the classical image denoising, where the total variation helps to obtain a smoother solution while preserving edges. 
Figure~\ref{fig:tv_image_examp} shows how the absolute spatial derivatives behaves for different inputs using input images (instead of intermediate layer features) as examples. It is observed that they provide different responses between clean and degraded images of the same scenes. 

\begin{figure}[bt]
\centering
\includegraphics[width=0.9\columnwidth]{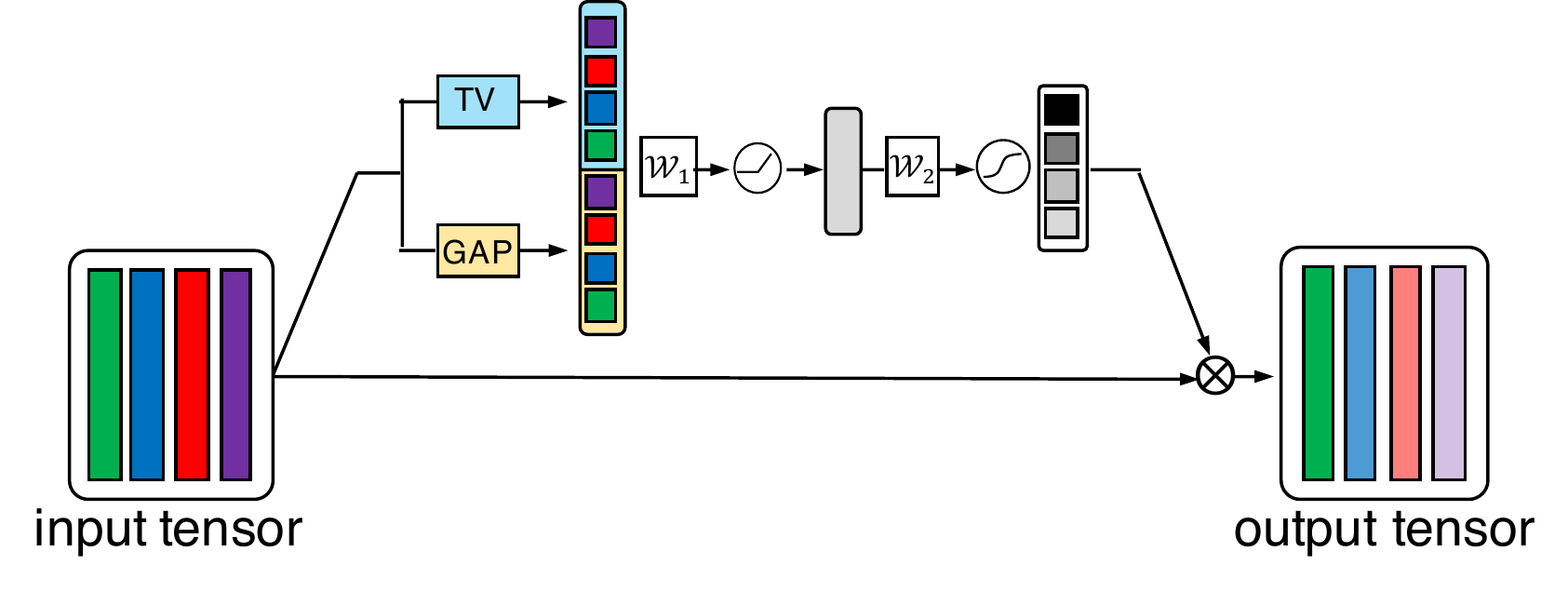}
\caption{The proposed attention mechanism improving the SE block. It generates channel-wise attention weights by global average pooling (the same as the standard SE block) and total variation (TV) of each channel activation values.}
\label{fig:tga}
%\vspace{-0.3cm}
\end{figure}

Figure~\ref{fig:tga} illustrates the proposed attention mechanism. We compute the global average and the total variation of activation values of each channel and input it to the same pipeline as the SE-block to generate attention weights over the channels. We replace the corresponding part in the SE-ResNet module with this reinforced attention mechanism. We will refer to the updated module as the ``improved SE-ResNet module'' in what follows.
% This mechanism is built into a ResNet module, as shown in Fig.~\ref{fig:res_tga}. 
We will show the effectiveness of the above design though experiments including ablation tests.

% \begin{figure}[t]
% \centering
% \includegraphics[width=0.9\columnwidth]{res_tga.pdf}
% \caption{The improved SE-ResNet Module, which incorporates the improved attention mechanism into a ResNet module.}
% \label{fig:res_tga}
% %\vspace{-0.3cm}
% \end{figure}

\subsubsection{Improved Design of a Dual Residual Block}
%{\color{red} [Intuition is written in line: 1004-1017]}}
\label{sec:fusion}
% figures
\begin{figure}[bt]
\centering
\includegraphics[width=0.9\columnwidth]{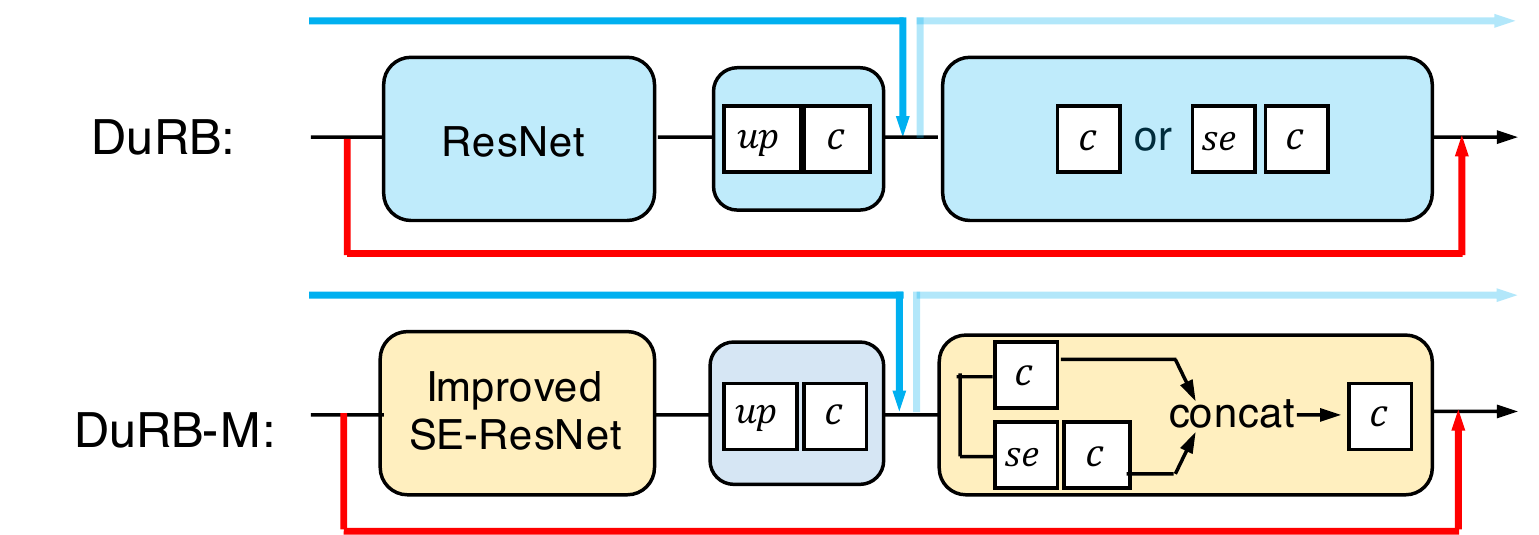}
\caption{The proposed basic block (DuRB-M) used for building our network.}
\label{fig:DuRB-M}
%\vspace{-0.3cm}
\end{figure}

The design of the dual residual networks \cite{DuRN} aims at making maximum use of paired operations that are believed to fit for image restoration tasks. The choice of the paired operations is arbitrary and four choices are suggested depending on the type of degradations. We pay attention on the two of them, in both of which the first operation is up-convolution. Specifically, one is the pair of up-convolution (i.e., up-sampling followed by convolution) and  simple convolution. The block employing the pair is named DuRB-U and applied to motion blur removal. (See the upper panel of Fig.~\ref{fig:DuRB-M}.) The other is the pair of up-sampling followed by convolution and a SE-ResNet module. The block is named DuRB-US and applied to haze removal. 

% \begin{table*}[t]
% \centering
% \caption{Results of an ablation test with different components and employment of multi-task learning.}
% %\vspace{0.3cm}
% \label{tab:ablation}
% \small
% %\resizebox{\columnwidth}{!}{ 
% \begin{tabular}{c|ccc|cccc|c} 
%   &TV& GAP& Fusion& motion blur & haze & rain-streak & JPEG artifact \\
%  \hline
%  &\xmark&\xmark&\xmark &  28.25 / 0.8724 & 26.58 / 0.9646 & 31.32 / 0.8976 & &\\
 
%  &\cmark&\xmark&\xmark &  28.49 / 0.8809 & 29.15 / 0.9699 & 31.99 / 0.9003 & &\\
 
%  &\xmark&\cmark&\xmark &  28.11 / 0.8703 & 29.66 / 0.9721 & 32.01 / 0.9015 & &\\
%  &\cmark&\cmark&\xmark &  28.51 / 0.8811 & 29.06 / 0.9719 & 32.03 / 0.9014 & &\\
%  \hline
%  MBN&\cmark&\cmark&\cmark &{\bf 28.91} / {\bf 0.8911} & {\bf 31.32} / {\bf 0.9778}& {\bf 32.15} / {\bf 0.9048}&\\
% \end{tabular}
% %}
% \end{table*}

In this paper, aiming at development of a block structure that can deal with multiple degradation factors
%motion blur, haze, and even more. Toward this end, 
we propose a new design, which we call DuRB-M. The idea is to integrate the above two designs (i.e., DuRB-U and -US) and also replace the ResNet module in the original DuRB structure with the aforementioned improved SE-ResNet module. To be specific, while keeping the same up-convolution for the first operation, we employ parallel computation of the second operations of DuRB-U and -US, i.e., convolution and a SE-block, for the second operation of the new block design; see the lower panel of Fig.~\ref{fig:DuRB-M}. The output maps of the two operations are merged by concatenation in the channel dimension, followed by $3\times 3$ convolution to adjust the number of channels. 
% We also 
% %{\bf [Xing: I changed SE-block to SE-ResNet module]} 
% with the enhanced attention mechanism, as shown in Fig.~\ref{fig:res_tga}.

\subsection{Design of the Output Branches}
% figure
\label{sec:outbranch}

\myparagraph{Minimal Design}
A minimal design of the output branches is to use a subnetwork of the same design but with different weights for each branch  specific to a single degradation factor. 
%Figure \ref{fig:alignment}(a) illustrates this design in the case of three tasks, 
Let $g_i$ $(i=1,\ldots)$ denote a subnetwork comprising one of the output branches. % and $f$ denotes the stem network.
% The basic design for the output branches is explained in this section. It has multiple decoders connected to the last layer of the stem net in parallel to each other. Figure~\ref{fig:MDRN}-(a) illustrates this design for three tasks, where $g_1$, $g_2$, and $g_3$ denote the decoders specified to the tasks, and $f$ denotes the stem-net consisting of five DuRB-M's. 
All $g_i$'s have an identical design, which starts with two sets of up-sampling plus convolution (implemented by PixelShuffle \cite{pixelshuffle}) and ReLU in this order, followed by convolution with a hyperbolic tangent activation function. The internal convolution layers all employ $3\times3$ kernels. The number of their channels are 96 for the first two conv. layers and 48 for the last one.

\myparagraph{Improved Design} %Multi-task Learning}
\label{sec:alignment_search}

Although the above design works fairly well, further  improvement can be achieved by inserting additional DuRB-M block(s), which is the same as those constituting the stem network, to some of the output branches. As shown in Fig.~\ref{fig:outermost}, the differences between different degradation factors are so large and they may not be fully absorbed by the lightweight $g_i$ of an identical design. We hypothesize that the difficulty with each degradation factor differs, and it can be handled by a hierarchical structure; easier factors are handled at a lower stage of the hierarchy and more difficult ones are at a higher stage. 
We conducted a systematic search, determining the following order for the primary four factors considered in this study: rain-streak, motion-blur, JPEG artifacts, and haze. To be specific, we stack three DuRB-M blocks in a row right after the stem network and insert $g_i$ ($i=1,2,3,4$) to each output of the stem network and the three stacked blocks; the inserted $g_i$'s handle the four factors in the above order.

\subsection{Design of Input Branches}

Although it is less essential to have multiple branches for the input of our network than for its output, we can achieve better performance by using different branches for the input. We adopt several ideas from the literature to design them. For motion blur, we employ the scale-recurrent \cite{tao-blur} and selective parameter sharing \cite{nest_blur} strategies, intending to restore clear textures progressively, and increase the capacity for removing various scales of blurring. We employ Gao \etal \cite{nest_blur}'s implementation for this branch and accordingly update $g_2$ in the input branch. For haze removal, the basic design seems powerful enough \cite{DuRN}. For JPEG-noise removal, we employ the same strategy used for motion blur removal. Further details are given in supplementary material.

\begin{table*}[t]
\centering
\caption{An ablation test to evaluate the effectiveness of several components in the proposed block, DuRB-M. A single model is trained jointly on the four factors. PSNR/SSIM is shown. }
%\vspace{0.3cm}
\label{tab:ablation}
\footnotesize
%\resizebox{\columnwidth}{!}{ 
\begin{tabular}{c|ccc|cccc|c} 
  &TV& GAP& Fusion& motion blur & haze & rain-streak & JPEG artifact \\
 \hline
 &\xmark&\xmark&\xmark &  28.60 / 0.8698 & 30.00 / 0.9789 &  32.37 / 0.9136 & 28.13 / 0.8254 & \\
 
 &\cmark&\xmark&\xmark & 28.92 / 0.8778 & 32.35 / 0.9815 & 32.65 / 0.9162 & 28.14 / 0.8253& \\
 &\xmark&\cmark&\xmark & 28.77 / 0.8737 & 32.48 / 0.9820 & 32.61 / 0.9150 & 28.14 / 0.8251& \\
 &\xmark&\xmark&\cmark & 29.32 / 0.8853 & 33.80 / 0.9846 & 32.78 / 0.9186 & 28.15 / 0.8268& \\
% &\cmark&\cmark&\xmark & 28.83 / 0.8751 & 32.34 / 0.9814 & 32.66 / 0.9159 & 28.12 / 0.8255& \\
 \hline
 MBN&\cmark&\cmark&\cmark &{\bf 29.35} / {\bf 0.8861} & {\bf 34.06} / {\bf 0.9860}& {\bf 32.83} / {\bf 0.9191}& {\bf 28.19} / {\bf 0.8271} & \\
\end{tabular}
%}
\end{table*}

\section{Training Method}

We train the proposed network in two steps. In the first step, we consider only its (non-recurrent) forward paths with different input-output branch pairs. We jointly train the forward paths on multiple degradation factors. After this step, the network can handle each degradation factor by using the corresponding input-output branch. In the second step, we fine-tune the network so that the network can deal with an image with unknown degradation factor by using the recurrent paths, as shown in Fig.~\ref{fig:face}. 

\subsection{Joint Training of Forward Paths}
\label{sec:joint}

We first train our network in different forward paths jointly on multiple factors in the following way. We split the training process into a series of cycles, in each of which we train the network on a combination of all the degradation factors. To be specific, each cycle contains one or more randomly chosen minibatches of a single factor. Considering that the loss decreases at a different speed for different factors, we choose the number of minibatches in one cycle, specifically, one for haze removal, one for rain-streak removal, one for JPEG artifacts removal, and three for motion blur removal. The minibatches are randomly chosen from the training split of each dataset and packed in a random order in a row for the cycle. We then iterate this cycle until convergence. Each input image in a minibatch is obtained by randomly cropping a region from an original training image. The crop size is 128$\times$128 pixels and 256$\times$256 pixels for rain-streak and the other factors, respectively. Further details are given in the supplementary material.

% \begin{figure}[t]
% \centering
% \includegraphics[width=.7\linewidth]{univ_train.pdf}
% \caption{{\color{red} univ. net training. Not good, the task-recurrent order changes. To be removed.} }
% \label{fig:univ_train}
% %\vspace{-0.3cm}
% \end{figure}

% \begin{table}[t]
% \centering
% \caption{{\color{red} Single task training vs. multi task training. Conducted on the simple version (single input branch) of our CNN.}}
% \label{tab:mul-vs-sin_final}
% \small
% %\resizebox{\columnwidth}{!}{ 
% \begin{tabular}{c|ccc|c} 
%       & blur & haze & rain &  \\
%  \hline
%   blur & 27.87 / 0.8653 & -/- & -/- & \\
%   haze & -/- & 28.58 / 0.9685 & -/- & \\
%   rain & -/- & -/- & 32.60 / 0.9139 & \\
%  \hline
%   Mult & 28.91 / 0.8911 & 31.32 / 0.9778 &  32.15 / 0.9048 & \\
% \end{tabular}
% %}
% \end{table}

\subsection{Finetuning to Enable Recurrent Computation}%Universal Network by Task-Recurrent Training}

After the joint training on the forward paths, we fine-tune the network so that it can be used in the recurrent computation mode. 
%train our network to make it a universal model for restoring images without knowing the distortion type for each image. 
We first choose in what order the multiple degradation factors are processed in the recurrent path. We conduct a test to choose their order, as it affects the final performance; details will be given in Sec.~\ref{sec:order}. We then train the network on this recurrent path in the following way. Choosing a training sample of a degradation factor, we input it into the recurrent path. We then compute the loss for the final output against the ideal clean image. As we know the degradation factor of the input, we also compute the loss for the intermediate output at the corresponding output branch against the same clean image. We then compute a weighted sum of the two losses with fixed weights (0.8 for the intermediate output and 0.2 for the last output in our experiments), minimizing it by backpropagation as usual. Using a larger weight on the intermediate output, we intend to make the forward path specializing in that degradation factor continue to remove it, while guaranteeing that the other paths do no harm on the final results. 

{
% We propose a task-recurrent training scheme for this purpose. Specifically, given a training sample with its distortion type known, we recurrently forward the input image through the network for all the tasks in a given order (discussed in Sec.\ref{}) of them. It is noteworthy that only the corresponding encoder and decoder, and the main-body of the network are used in each forward. We compute loss at the decoder corresponding to the sample's distortion type, and the decoder used in the last forward. We combine the two losses with a pair of weights (0.8 to the former, 0.2 to the later) summing to 1. Assigning a larger/smaller weight to the former/later loss, we intend to direct the network to solve an individual task in its corresponding forward, while keeping the output from being twisted in the later forwards as possible. Note that when the two decoders are identical, the combined loss is equivalent to the loss computed with weight$=1$. 

%\subsection{Orders for Task-Recurrent Training}
% We search for the optimal order of tasks for executing our network. We apply the same strategy as we do so for searching decoders' alignments (see Sec. \ref{sec:alignment_search}). Specifically, we try out all six possible orders of the three tasks (\ie motion blur, haze, rain-streak removal) to find the one performing best under PSNR. Then, we ``insert'' the fourth task (\ie jpeg artifacts removal) into the optimal order found for the three tasks. We found through experiments that J$\rightarrow$H$\rightarrow$B$\rightarrow$R is the best order.
}
\section{Experimental Results}
% We conduct experiments to evaluate the proposed method. We choose three tasks, i.e., motion blur removal, haze removal, and rain-streak removal, for the main experiments (i.e., detailed architectural design search, ablation study, etc.); we additionally use JPEG compression noise removal for performance evaluation. 

%\subsection{Experimental Configuration}

\subsection{Ablation Test on the DuRB-M Block}
\label{sec:apples_to_apples}

We have proposed two design improvements of the Dual Residual Block for multi-task learning. We present here an ablation study to 
%on the benchmark datasets (introduced in Sec.\ref{sec:perf_benchmark})} that 
evaluate the contribution of each design component. For this purpose, we train a number of networks with different configurations according to the training step of Sec.~\ref{sec:joint} on the same datasets of four degradation factors as Sec.~\ref{sec:perf_benchmark}. The stem net consists of five stacks of the DuRB-M block. For the sake of efficiency, we chose the minimal design for the output branches (Sec.~\ref{sec:outbranch}) and  identical design for the input branches (thus, there is only a single input branch). See the supplementary material for details.
Table \ref{tab:ablation} shows the results. 
TV and GAP mean the channel-wise Total Variation and the channel-wise Global Average Pooling, respectively, both of which are used for attention computation. Fusion means the improved design of the dual residual block that employ fused operations (Sec.~\ref{sec:fusion}). It can be confirmed that the use of all the three components yields the maximum accuracy for all the factors. 
%It is also observed that each component has a certain amount of positive impact on the resulting accuracy, although it differs for different degradation types. 

\subsection{Order of Factors in the Recurrent Path}
\label{sec:order}

In this study, we consider four degradation factors. Thus, the proposed network has four pairs of input-output branches. To examine the impact of the order of these factors processed in the recurrent path, we conducted an experiment. As exhaustive search needs tests on 24 order patterns, we employ a two-step approach. First, we choose three factors and test all the order patterns for them. Then, for the best order of the three factors, we insert the last factor into one of the four places. Table~\ref{tab:task-recurr} shows the results; the first test on the three chosen factors, i.e., motion-blur, haze, and rain-streak is shown on the left and the second test on the last factor, i.e., JPEG noise on the right. It is seen that at least the orders of the first three factors have non-negligible effects. From the results, we choose the order J$\rightarrow$H$\rightarrow$B$\rightarrow$R for the subsequent experiments. 
\begin{table}[t]
\centering
\caption{Order of the degradation factors processed in the recurrent path. The results are averaged over three/four tasks.}
\label{tab:task-recurr}
\small
\resizebox{\columnwidth}{!}{ 
\begin{tabular}{cc}
\begin{tabular}{c|c|c|c} 
      & order & PSNR / SSIM&   \\
 \hline
    & B$\rightarrow$H$\rightarrow$R& 31.83 / 0.9396& \\
    & B$\rightarrow$R$\rightarrow$H& 31.84 / 0.9395& \\
    & {\bf H$\rightarrow$B$\rightarrow$R}& 32.07 / 0.9403& \\
    & H$\rightarrow$R$\rightarrow$B& 32.06 / 0.9402& \\
    & R$\rightarrow$B$\rightarrow$H& 31.81 / 0.9128& \\
    & R$\rightarrow$H$\rightarrow$B& 32.02 / 0.9401& \\
 \hline
    & & & \\
 \end{tabular}
 \begin{tabular}{c|c|c|c} 
  & order & PSNR / SSIM& \\
  \hline
    & {\bf J$\rightarrow$H$\rightarrow$B$\rightarrow$R} &31.14 / 0.9123 & \\
    & H$\rightarrow$J$\rightarrow$B$\rightarrow$R &31.13 / 0.9121 & \\
    & H$\rightarrow$B$\rightarrow$J$\rightarrow$R &31.12 / 0.9121 & \\
    & H$\rightarrow$B$\rightarrow$R$\rightarrow$J &31.13 / 0.9121 & \\
  \hline
    & & & \\
\end{tabular}
\end{tabular}
}
\end{table}

\begin{table*}[t]
\centering
\caption{Comparison of the proposed method (MBN/R-MBN) with state-of-the-art methods  on the removal task of each of four degradation factors (PSNR/SSIM). Methods in the ``dedicated'' block are designed for a single degradation factor. MBN is trained on all the factors but requires specification of the target factor. R-MBN and cascade do not need it. The best one is in bold and the second is with underline. 
% The value with the superscript $^*$ is considered to be an error.
%The {\bf best} and the \underline{second best} result. 
%A method with a superscript $^{*}$ means its result is unreplicable by us.
}
\label{tab:multi-task-res}
\resizebox{\textwidth}{!}{
\begin{tabular}{c|cc||cc||cc||cc|c}
 baselines& \multicolumn{2}{c||}{motion blur removal} & \multicolumn{2}{c||}{haze removal} & \multicolumn{2}{c||}{rain-streak removal}& \multicolumn{2}{c|}{JPEG artifacts removal (q=10)} & \\

 \hline
 \multirow{4}{*}{dedicated} & Liu \etal \cite{DuRN} & 29.90 / 0.9100 & Cai \etal \cite{dehazenet}  & 19.82 / 0.82 & Wang \etal \cite{SPANet} & 30.05 / {\bf 0.93} & Dong \etal \cite{ARCNN_jpeg}& 28.98 / 0.82 & \\

 &  Zhang \etal \cite{patch_blur} & 30.63 / 0.9053 & Ren \etal \cite{GFN} &  24.91 / \underline{0.92}  & Li \etal \cite{rescan} & 32.48 / 0.91 & Chen \etal \cite{TNRD_jpeg} & 29.15 / 0.81 & \\
 
 &  Purohit \etal \cite{moment_blur} & \underline{30.79} / 0.9100 & Liu \etal \cite{DuRN}  & {32.12} / {\bf 0.98}  & Li \etal \cite{derain_sota_acm} & \underline{33.16} / \underline{0.92} & Zhang \etal \cite{dncnn} & \underline{29.19} / 0.81& \\
 
 & Gao \etal \cite{nest_blur} & {{\bf 31.35} / {\bf 0.9174}}  & Liu \etal \cite{iccv19_haze} &  32.16 / {\bf 0.98}  &  Liu \etal \cite{DuRN} & {33.21} / {\bf 0.93} & Zhang \etal \cite{nonlocal_iclr} & {\bf 29.63} / \underline{0.82} & \\
 \hline

 MBN & \multicolumn{2}{c}{30.70 / \underline{0.9111}} & \multicolumn{2}{c}{{\bf 32.68} / {\bf 0.98}} &\multicolumn{2}{c}{{\bf 33.40} / {\bf 0.93} }& \multicolumn{2}{c|}{28.24 / {\bf 0.83}}& \\
\hline \hline

cascade & \multicolumn{2}{c}{17.48 / 0.7306} & \multicolumn{2}{c}{27.03 / 0.87}& \multicolumn{2}{c}{21.52 / 0.76}& \multicolumn{2}{c|}{16.97 / 0.64}& \\

R-MBN & \multicolumn{2}{c}{30.67 / 0.9110} & \multicolumn{2}{c}{\underline{32.38} / {\bf 0.98}} &\multicolumn{2}{c}{{\bf 33.40} / {\bf 0.93}} & \multicolumn{2}{c|}{28.19 / {\bf 0.83}}& \\

%  &\multicolumn{2}{c}{-/-\cite{suganuma_cvpr19_arxiv}}& \multicolumn{2}{c}{-/-}& \multicolumn{2}{c}{-/-}& \multicolumn{2}{c|}{-/-}& \\

\end{tabular}
}
\end{table*}

\begin{table*}[bt]
\centering
\caption{Results of restoration from images with mixed degradation factors. R-MBN is trained only on images with a pure, uncombined factor. R-MBN* and Suganuma et al. \cite{suganuma_cvpr19_arxiv} are trained on images with mixed factors. }
% Results comparison to show generalization ability of the proposed method. {\color{red} [MBN-chain means the model is trained on multiple single tasks at first, and then finetuned using the unknown combined data which is used to train Suganuma \etal's model.]  }}
\label{tab:new_data_results}
\footnotesize
%\resizebox{\columnwidth}{!}{ 
\begin{tabular}{c|cccc|c|cc|c} 
  &rain& blur& haze& JPEG& R-MBN& R-MBN$^{*}$&Suganuma \etal \cite{suganuma_cvpr19_arxiv}\\
 \hline
 \multirow{4}{*}{One task}
 &\cmark&\xmark&\xmark &\xmark& {\bf 38.63} / {\bf 0.9797} & 34.84 / 0.9684 & 30.50 / 0.8660 & \\
 &\xmark&\cmark&\xmark &\xmark& {\bf 33.66} / {\bf 0.9525} & 31.41 / 0.9429 & 27.55 / 0.8411 & \\
 &\xmark&\xmark&\cmark &\xmark& {\bf 30.07} / {\bf 0.9710} & 30.61 / 0.9747 & 25.73 / 0.9019 & \\
 &\xmark&\xmark&\xmark &\cmark& {\bf 31.72} / {\bf 0.9154} & 30.89 / 0.9127 & 28.99 / 0.8584 & \\
 \hline
 \multirow{6}{*}{Two tasks}
 &\cmark&\cmark&\xmark &\xmark& 28.59 / 0.8454 & {\bf 29.91} / {\bf 0.9072} & 26.13 / 0.7853 & \\
 &\cmark&\xmark&\cmark &\xmark& {\bf 28.42} / 0.9441 & 28.41 / {\bf 0.9451} & 24.12 / 0.8139 & \\
 &\cmark&\xmark&\xmark &\cmark& 25.66 / 0.7515 & {\bf 29.25} / {\bf 0.8753} & 27.18 / 0.7828 & \\
 &\xmark&\cmark&\cmark &\xmark& 26.58 / 0.9163 & {\bf 27.57} / {\bf 0.9216} & 23.27 / 0.7776 & \\
 &\xmark&\cmark&\xmark &\cmark& 26.90 / 0.8012 & {\bf 27.42} / {\bf 0.8292} & 26.04 / 0.7714 & \\
 &\xmark&\xmark&\cmark &\cmark& 24.89 / 0.8548 & {\bf 26.60} / {\bf 0.8693} & 24.42 / 0.8075 & \\
 \hline
 \multirow{4}{*}{Three tasks}
 &\cmark&\cmark&\cmark &\xmark& 22.57 / 0.7894 & {\bf 26.12} / {\bf 0.8786} & 22.38 / 0.7195 & \\
 &\cmark&\cmark&\xmark &\cmark& 23.64 / 0.6912 & {\bf 26.20} / {\bf 0.7879} & 24.91 / 0.7214 & \\
 &\cmark&\xmark&\cmark &\cmark& 21.15 / 0.7045 & {\bf 25.23} / {\bf 0.8269} & 22.04 / 0.6732 & \\
 &\xmark&\cmark&\cmark &\cmark& 21.29 / 0.7154 & {\bf 24.71} / {\bf 0.7843} & 22.59 / 0.7117 & \\
 \hline
 \multirow{2}{*}{Four tasks}
 &\cmark&\cmark&\cmark &\cmark& 18.84 / 0.6249 & {\bf 23.58} / {\bf 0.7436} & 22.04 / 0.6733 & \\
 &\multicolumn{4}{c|}{Random combinations}& 25.41 / 0.7881 & {\bf 27.37} / {\bf 0.8502} & 24.58 / 0.7618 & \\
 \hline
 %&\multicolumn{4}{c|}{} & & & \\
\end{tabular}
%}
\end{table*}

\begin{figure*}[t]
\centering
\includegraphics[width=\textwidth]{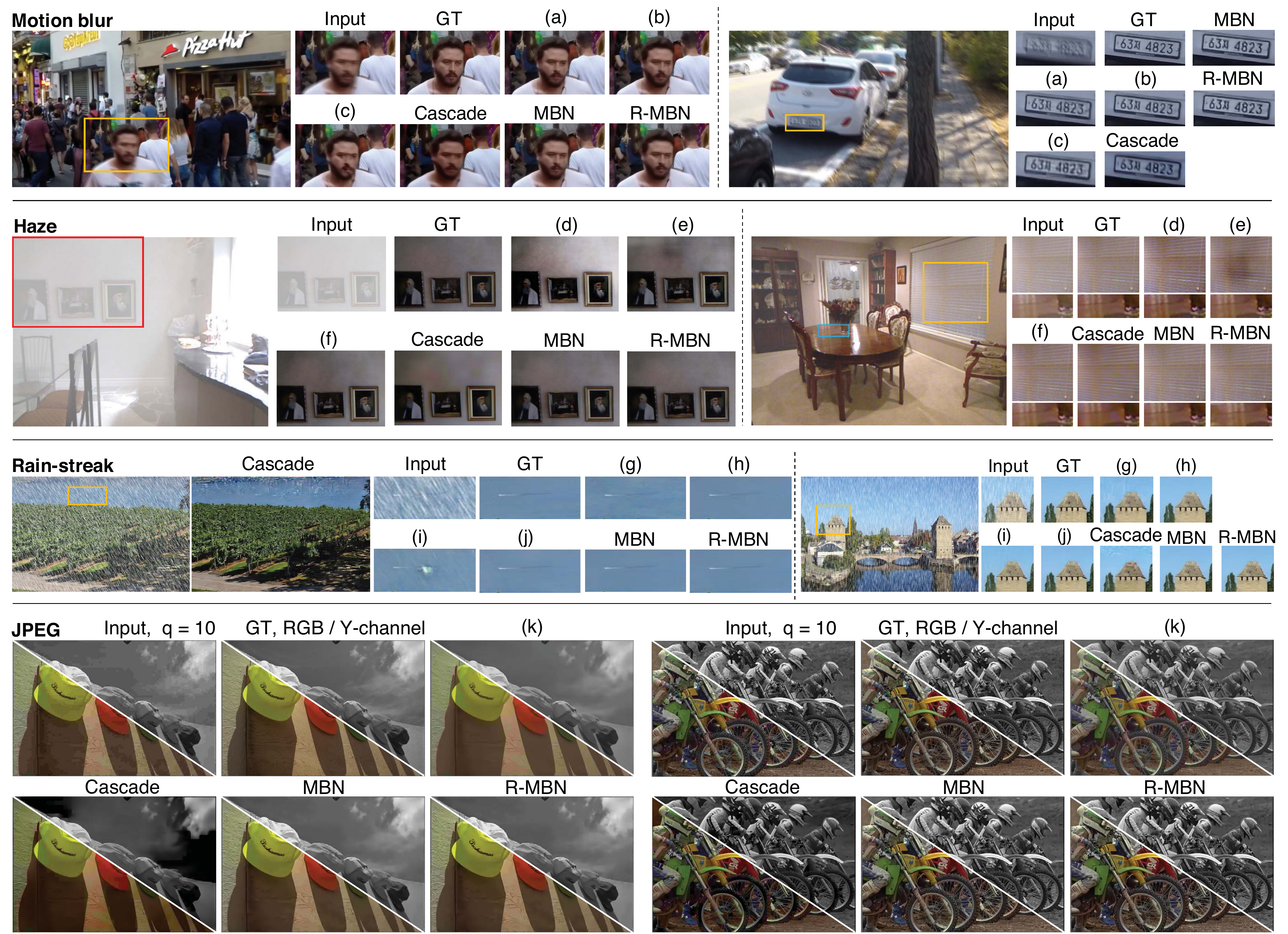}
\caption{Qualitative comparison on the four degradation factors. Motion blur: (a) \cite{patch_blur}; (b) \cite{nest_blur}; (c) \cite{DuRN}. Haze: (d) \cite{GFN}; (e) \cite{iccv19_haze}; (f) \cite{DuRN}. Rain-streak: (g) \cite{rescan}; (h) \cite{derain_sota_acm}; (i) \cite{umrl}; (j) \cite{DuRN}. JPEG artifact: (k) \cite{nonlocal_iclr}. ``Cascade'' is the cascade of four models as in Table~\ref{tab:multi-task-res}.}
\label{fig:four_vis}
%\vspace{-0.3cm}
\end{figure*}

\subsection{Performance on Single Factor Removal}
\label{sec:perf_benchmark}
%We compare the proposed method with the state-of-the-art methods for different tasks. Table \ref{tab:multi-task-res} shows the results. We choose the best four published methods (ranked by PSNR) for each task. ``Ours(3)'' indicate our method trained on the three tasks. We report here the accuracy values obtained for the best architecture found in the experiment explained above. It is observed that the proposed method outperforms others for haze removal and achieves comparable performance to the previous methods for other tasks. 

We first show the performance of our models on single factor removal tasks. We consider two models. One is the model obtained right after the first training step, which has multiple input-output pairs. The other is the final model obtained after the fine-tuning, which has a single input and output.
%In this test, we input an image with a single degradation factor to the input of our model. Using its internal recurrent path, our model yields a restored image from its output, without using the knowledge of the factor.
Their difference is whether we need to specify the degradation factor of the input images or not. 
%We compare these two with the state-of-the-art methods designed and trained for a particular factor on the removal task of the same factor.

%\vspace*{-3mm}
\myparagraph{Datasets} 
We choose dataset(s) for each task that is the most widely used in recent studies. We use the GoPro dataset \cite{nah} for motion blur removal. It consists of 2,013 and 1,111 non-overlapped training (GoPro-train) and test (GoPro-test) pairs of blurred and sharp images, respectively. We use the RESIDE dataset \cite{RESIDE} for haze removal, which consists of 13,990 samples of indoor scenes and a few test subsets. Following \cite{DuRN} and \cite{GFN}, we use a subset SOTS (Synthetic Objective Testing Set) that contains 500 indoor scene samples for evaluation. 
% For both training and test subsets of RESIDE, synthetic hazy effects are made using \eqref{equ:haze_model}.
We use the DID-MDN dataset \cite{derain_zhanghe} for rain-streak removal. It consists of 12,000 training pairs and 1,200 test pairs of clear image and synthetic rainy image. 
%Rain-streak effects for an image in this dataset are made using Photoshop. 
As for JPEG artifacts removal, we use the training subset (800 images) of the DIV2K dataset \cite{div2k} and LIVE1 dataset (29 images) for training and testing our networks, following the study of the state-of-the-art method \cite{nonlocal_iclr}. 
% An additional setting we made for our experiment is that we re-sized the original DIV2K images (larger than $2000 \times 1000$ for most of them) to their half for training out network for computational efficiency.

%\vspace*{-3mm}
\myparagraph{Results} 
Table \ref{tab:multi-task-res} shows the results. It includes the performance of four best published methods for each degradation factor (``dedicated''). We use the authors' codes for them. MBN indicates our model after the 1st training step and R-MBN indicates our model after the fine-tuning. As a baseline of the latter, we consider a cascade of the four best dedicated models (i.e., the last row in the ``dedicated'' block), which at least formally does not need specification of a degradation factor. To be specific, we cascade the models of Zhang \etal \cite{nonlocal_iclr}'s, Liu \etal \cite{DuRN}'s, Gao \etal \cite{nest_blur}'s, and Liu \etal \cite{DuRN} in this order (J$\rightarrow$H$\rightarrow$B$\rightarrow$R). 
% We compare our models with two types of baselines, which are i) the state-of-the-art methods that are dedicated to each single tasks (``dedicated''), and ii) an image restoration pipeline that cascades the best (ranked in PSNR) dedicated model of each task for the four tasks (``cascade''). 
%We have an exception with motion blur removal in ii), we use zhang \etal \cite{}'s model (the second best) instead of Purohit \etal \cite{}'s for their code which is not publicly available.
% {\color{red} (To be shorten) We make another statement that there are some recently published studies reporting outstandingly high results in SSIM compare to other baselines for motion blur removal. However, their code for computing SSIM is blind. For the sake of a fair comparison, we only include those for which either trained model or de-blurred results are available, into our table.} 

It is seen from the table that MBN yields at least comparable performance to the best dedicated models. More importantly, R-MBN, which has only a single input and output and thus can deal with images with any degradation factor, achieves very similar performance to MBN. This will be a big advantage in real world applications, despite some gap with the best dedicated model (e.g., motion blur). Figure \ref{fig:four_vis} shows a few examples of restoration from each of the four factors. It can be observed that the proposed MBN can restore high-quality images comparable to the state-of-the-art methods, and R-MBN attains similar quality to MBN. These agree well with the above quantitative comparison. It is also seen that the cascade of the four networks yield only sub-optimal results. For JPEG, we show RGB images as well as Y-channel images, as the latter are mainly used in the previous studies including RNAN \cite{nonlocal_iclr}. It is seen from the restored RGB images that our networks recover more precise colors.

% Table \ref{tab:multi-task-res} shows the results obtained by simultaneously training our network on four tasks, i.e., the three tasks plus JPEG compression noise removal. It is seen that the addition of this task contributes to further improvements on haze removal and rain-streak removal. In the experiment, we search for a good design for the four tasks; to do this with a modest computational cost, we considered only combinations of inserting either a new decorder or a new decoder with a DuRB-M to the above three-task network. 
% %{\color{red} And which one is the best, which I think yielded the reported numbers in the table?} 
% The best performer is the one with additional DuRB-M inserted in between B and H, i.e.,  R$\rightarrow$B$\rightarrow$J$\rightarrow$H, where $J$ is the JPEG compression noise removal. The results with ``Ours(4)'' in Table~.\ref{tab:multi-task-res} is obtained by this design.
% A few examples of the output images for the four tasks are shown in Fig.~\ref{fig:four_vis}. 

\subsection{Removal of Unknown Combined Degradations}

An even harder task is to restore images with mixed degradation factors with unknown mixing ratios. We tested two versions of our model on this task. One is the model trained by the aforementioned method on training images having a single degradation factor, i.e., R-MBN in Table \ref{tab:multi-task-res}. We also train the same network on a set of images with mixed degradation factors, which are generated as explained below. We refer to it as ``R-MBN*''. We compare these with the method of Suganuma \etal \cite{suganuma_cvpr19_arxiv}, for which, following their paper, we train their network on the same set of images with mixed degradation factors. We omit the results of the earlier study \cite{toolchain} on this task, since they perform worse than  \cite{suganuma_cvpr19_arxiv}, as is reported in \cite{suganuma_cvpr19_arxiv}.

% {\color{red} Yu \etal \cite{toolchain} and Suganuma \etal \cite{suganuma_cvpr19_arxiv} studied this task. }

% restore clear images from mixed degradation types with unknown mixing ratios. Yu \etal start this research trend with a framework that multiple light-weight CNNs are trained for different image distortions and are adaptively applied to input images by a mechanism learned by deep reinforcement learning. However, they only deal with combinations of laboratory level distortions (\ie Gaussian noise, Gaussian blur and Jpeg artifacts). Suganuma \etal proposed an attention-based operators selecting strategy to handle this problem, showing performance better than Yu \etal's. Moreover, they also addressed more realistic distortion types such as motion blur and raindrop. 

\myparagraph{Dataset} 
%Towards comparing our approach with Suganuma \etal's, 
% Towards analysing how good can our approach generalize to combined distortions, 
Since the datasets used in the previous studies \cite{toolchain,god_suganuma} are not fit for the purpose here, we created a new dataset covering the same four degradation factors. For this purpose, we use the outdoor scenes in RESIDE-$\beta$ \cite{RESIDE}, which is the dataset for haze removal, as source images and add synthesized degradation to them. 
% we propose a novel dataset of single and combined distortion of motion blur, haze, rain-streak and JPEG artifacts. 
% We employ the outdoor scenes in RESIDE-$\beta$ \cite{RESIDE} as our source images. 
We synthesize motion blur effects, rain-streak effects, and JPEG artifacts by the method of Hendrycks \etal \cite{distort_imagenet}, GIMP  on Ubuntu 18.04.2 \cite{gimp}, and the Python Imaging Library, respectively. 
% Rain-streak effects are synthesize on the source images by GIMP \cite{gimp} on Ubuntu 18.04.2. 
For haze, we use the original data in the RESIDE-$\beta$ dataset.
% JPEG artifacts are made for the source images using the Python Imaging Library.
Then, we create two training datasets; a set of images with a pure degradation factor out of the four and a set of images with combined degradation factors. For the latter, we create each image so that it has a combination of $X$ factors, where $X$ is a random number chosen from $\{1,2,3,4\}$. We create two groups of test sets associated with the two training datasets. The first group contains four test sets of images with a pure degradation factor and the second group contains twelve test sets covering all the combinations of either two, three, or four factors; see Table \ref{tab:new_data_results}. Further details are given in the supplementary material. 
% For the combined distortion effects, we randomly choose a number(from [1,2,3,4]) of distortions to be added onto the source images to make the training set. For the test sets, we combine two/three/four/random number(from [1,2,3,4]) of factors, yielding 12 test sets (see Table.~\ref{tab:new_data_results}).
% We decide the strongness of a distortion for all the processes mentioned above by randomly selecting the corresponding parameters from their ranges. Details about generating the dataset are given in the supplementary material. 
% In summary, the proposed dataset contains four groups of training/test subsets for single tasks, and a training set with 12 test sets for mult-task. 
% % they are used to train/evaluate models for motion-blur, haze, rain-streak and jpeg artifacts, and combined distortion removal.

\myparagraph{Experiments} 
%We compare the performances of our method and the method of Suganuma \etal \cite{} on the above dataset. 
We consider two versions of our model trained differently; they have the same architecture as the last experiment. The first model is trained on the training set of pure, uncombined factors by the aforementioned two-step method. The second model is obtained by fine-tuning the first one from its first step's checkpoint on the training set of combined factors. The method of Suganuma \etal is also trained on them, following the suggested method in their paper. Then we test the three models on all the test sets, i.e., the four of pure factors and the twelve of combined factors. 

% The proposed R-MBN ({\color{red} to be changed}) is trained in two manners; i) we trained it only using the four single distortion training subsets. ii) we finetune the model trained in i) using the combined distortion training subset. We denote the model trained with i) and ii) by R-MBN and R-MBN$^{*}$, which is used in Table~\ref{tab:new_data_results}.
% For the baseline method (Suganuma \etal \cite{suganuma_cvpr19_arxiv}), we train it using the combined distortion training subset, following the original instructions provided with their paper. 
% We evaluate the three models using all 12 test subsets to analyse how good/bad the proposed model generalizes to this task.

% We train the proposed models (R-MBN) using the four single distortion training subsets, and evaluate it on both the single distortion test subsets and the combined distortion test subsets. For Suganuma \etal's network, we train it using the combined distortion training subset, and evaluate it on the same (all) test sets. Note that it is more difficult for our CNNs to achieve a good level of performance on the combined distortion test set than theirs. In training, Our CNNs do not view combined distortions while theirs views both single and combined distortions. More details are given in the supplementary material.

\myparagraph{Results} Table \ref{tab:new_data_results} shows the results. 
It is seen that unsurprisingly, our model trained on single-factor data (R-MBN) achieves the best performance on single-factor test sets; and that trained on combined-factor data (R-MBN*) achieves the best performance on combined-factor test sets. More importantly, the latter outperforms the method of Suganuma \etal with large margins, while it is not so inferior to the model trained on pure factors (R-MBN). Thus, from a practical application point of view,  R-MBN* will be the first choice for images having combined factors and R-MBN is recommended for images with a single factor. Additionally, although R-MBN is inferior to R-MBN* in the case of combined factors, the gap is not so large up to combination of two factors, and it is still mostly better than the method of Suganuma et al. Considering the fact that R-MBN is trained only on images with pure factors, this result is encouraging for a further study toward ``universal image restoration method'' that can handle any degradation factors including those that have not seen before. 

% that the proposed method outperforms the baseline method by a large margin for single task evaluation. The differences decreases with increasing number of distortion factors combined for an image....

\section{Conclusion}

In this paper, we have considered the problem of restoring a clean image from its degraded version with an unknown degradation factor, subject to the condition that it is one of the known factors. To solve this problem, we proposed a network having multiple pairs of input and output branches and use it in a recurrent fashion such that a different branch pair is used at each of the recurrent paths. The shared part of the network is build upon dual residual networks recently proposed by Liu et al. We showed improved designs of its component block, named DuRB-M, that enables to handle different degradation factors. We have also proposed a two-step training method for the network, which consists of multi-task learning and fine-tuning. We showed through several experimental results the effectiveness of the proposed method. 

{\small
\bibliographystyle{ieee_fullname}
\bibliography{egbib}
}

\clearpage
\setcounter{figure}{6}
\setcounter{table}{4}
\appendix

\begin{figure*}[bt]
\centering
\includegraphics[width=.95\textwidth]{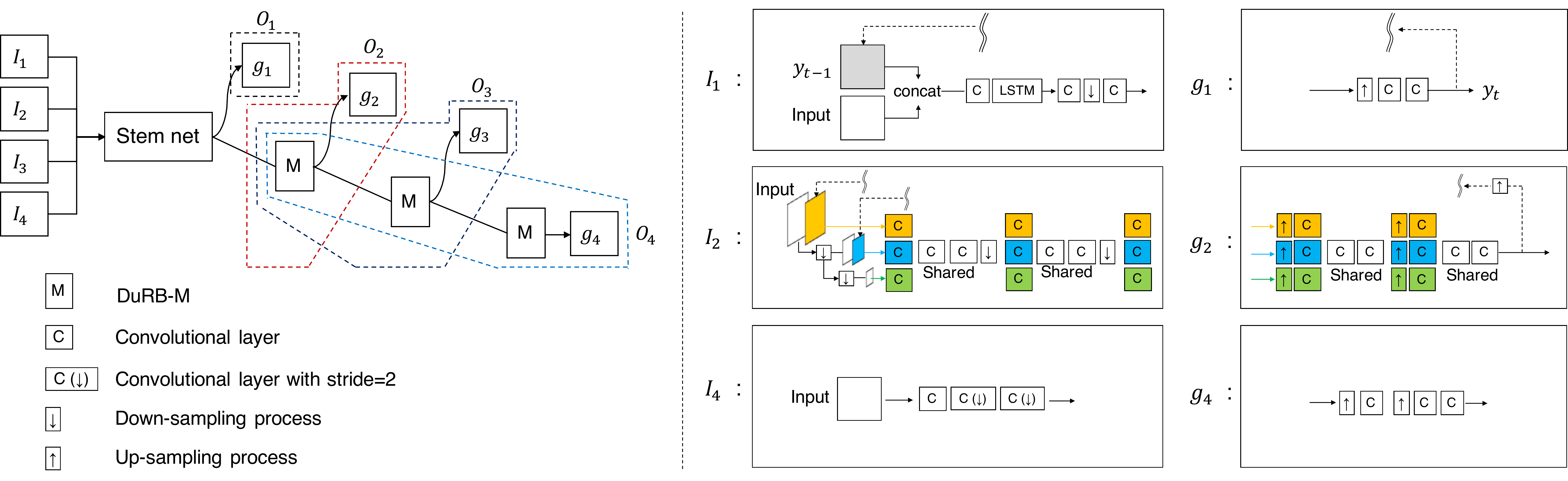}
\caption{An illustration of the architecture of the proposed network with the optimized input/output branches. ($I_1,O_1$) is used for rain-streak. ($I_2,O_2$) is for motion-blur. ($I_3,O_3$) is for JPEG artifacts. $(I_4,O_4)$ is for haze. }
%Ok. $I_{3}$ and $O_{3}$ are designed same as $I_{2}$ and $O_{2}$. }
\label{fig:RMBN}
\end{figure*}

\section*{Appendix}
\section{Detailed Design of Input/Output Branches}

The input-output branch pairs comprising the proposed network can have different designs for different degradation factors, as they are specifically used for an individual factor. The network works fairly well even when employing the same design (but with unshared weights) for all the factors considered in this study. However, as explained in the main paper,  the network with factor-specific input/output branches achieves the best performance; the results reported in  Sec.5.3 and 5.4 were obtained by one such design. In the main paper, we provide only a brief summary about it due to the lack of space. Here, we provide its details as well as underlying thoughts and experiments conducted to validate its design. 
% We introduce the designs for each input branch and its updated output branch in this section.

\subsection{Hierarchical Design of Output Branches}

\begin{figure}[t]
\centering
\includegraphics[width=.85\columnwidth]{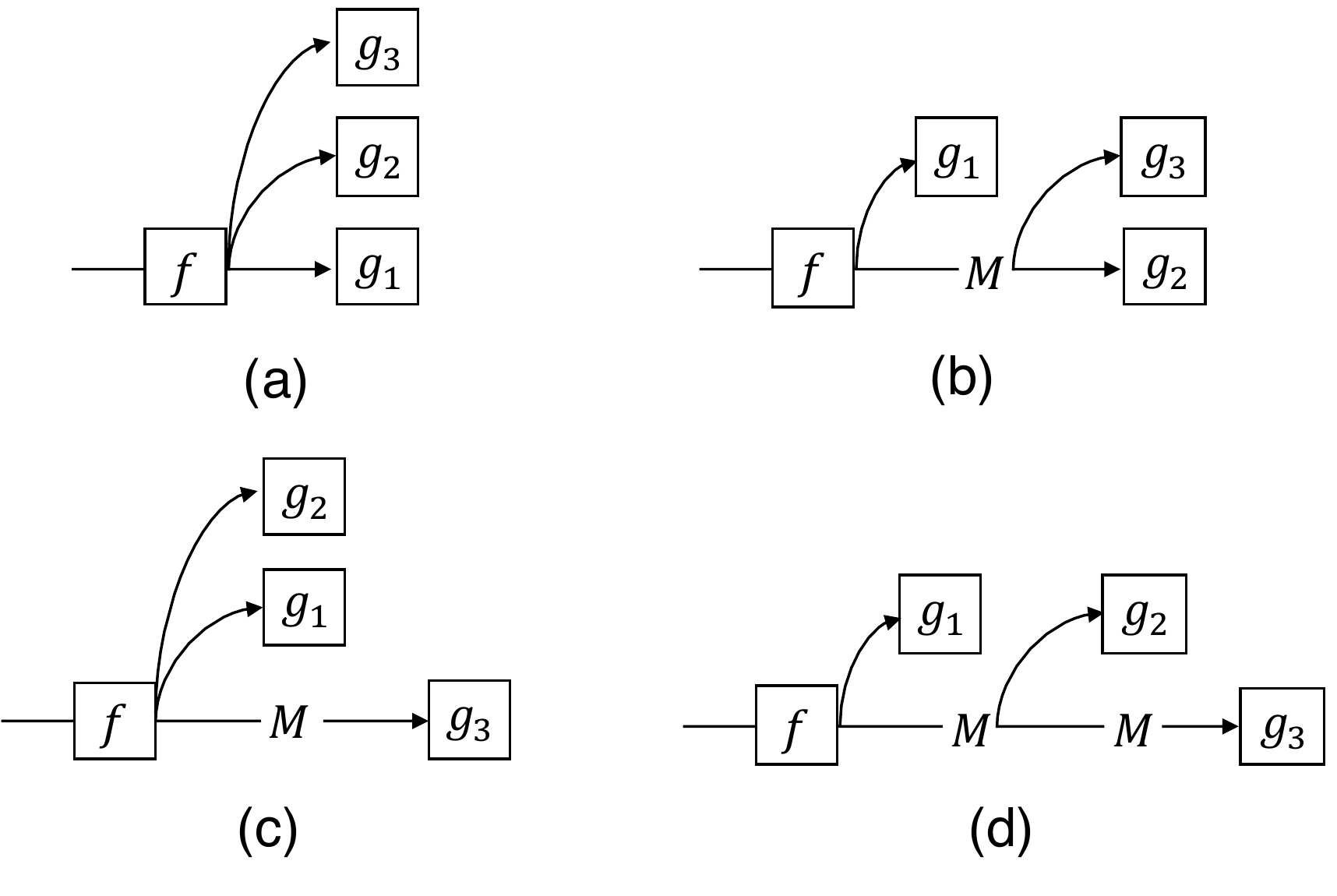}
\caption{Possible configurations of output branches with additional DuRB-M blocks. M indicates a DuRB-M block. Note that we consider configurations of selected three factors in our comparative experiments for the sake of efficiency.}
\label{fig:alignment}
\end{figure}

Figure \ref{fig:RMBN} shows the the proposed network with the complete illustration of all the branches.
As briefly explained in Sec.~3.3, we employ a hierarchical design utilizing additional DuRB-M blocks in the design of output branches for different factors, as shown in the upper-left of Fig.~\ref{fig:RMBN}. 
%{\color{red} %Fig.~\ref{fig:alignment}}. 
We found this design to performs the best in an experiment, where we compared possible configurations of three factors shown Fig.~\ref{fig:alignment}; we chose the three factors ( motion blur, haze, and rain-streak) for the sake of efficiency. 

Assuming the above result applies to the case of four factors, we ran a systematic search to find the best alignment of $g_1,\ldots,g_4$ and the four factors. As there are 24 ways of alignment in total, we employ the following two-step search strategy for the sake of efficiency. We first search for the optimal alignment for three selected factors, motion blur, haze, and rain-streak. We then consider inserting JPEG artifacts (J) into the best alignment of the three factors.
%We set stride$=2$ for the second and the third layer for down-sampling an input image. % order for inserting $g_{i}$ ($i=1,2,3,4$) (Sec. 3.3 in the main-text). For the sake of efficiency, We configure our network to have an identical input branch and a stack of five DuRB-M's for the stem net. The input branch consists of For
To make this search efficient, we also employ an identical design for the input branches as well as output branches for all the factors. The former consists of three conv. layers with $48$, $96$, and $96$ channels and with stride $=1$, 2, and 2, respectively. For the latter, we use the minimal design introduced in Sec.~3.2 for all $g_i$'s ($i=1,2,3,4$).

Table \ref{tab:output_branch3} shows the results of the first stage of this search. `$\rightarrow$' indicates a DuRB-M block. It is seen that R$\rightarrow$B$\rightarrow$H performs the best.  We then test inserting J into R$\rightarrow$B$\rightarrow$H, as shown in Table \ref{tab:output_branch4}. 
%It is seen in the table that 
%i) all the candidates perform similarly good in SSIM,
%ii) R$\rightarrow$J$\rightarrow$B$\rightarrow$H, R$\rightarrow$B$\rightarrow$J$\rightarrow$H and JR$\rightarrow$B$\rightarrow$H perform similarly good, better than the others in PSNR. 
Based on the results, we choose R$\rightarrow$B$\rightarrow$J$\rightarrow$H. 
%It is noteworthy that JR$\rightarrow$B$\rightarrow$H denotes the configure that the output branches of J and R are inserted into the same stage, \ie no DuRB-M between them; while there are two DuRB-M's used between R and B, B and H respectively. Training details are given in Sec. \ref{sec:train_details}
Note that this `order' of the factors in the network design is different from the order of the degradation factors processed in a recurrent path, which is chosen to be J$\rightarrow$H$\rightarrow$B$\rightarrow$R, as is explained in Sec.~5.2; that is, J is first removed by using the corresponding input/output pair, next H is  removed by using the corresponding pair, similarly, B and R. 

%\section{Systematic Search for Hierarchy of Factors in the Design of Output Branches}
\label{sec:decoder_algin}

\begin{table}[t]
\centering
\caption{B: motion-blur removal, H: haze removal, and R: rain-streak removal. The values (PSRN/SSIM are averaged accuracy over the three tasks.}
\label{tab:output_branch3}
\resizebox{\columnwidth}{!}{
\begin{tabular}{cc}
\begin{tabular}{c|c|c|c}
 & Alignment & PSNR / SSIM & \\
 \hline
 & B$\rightarrow$H$\rightarrow$R & 30.61 / 0.9244& \\
 & B$\rightarrow$R$\rightarrow$H & 30.42 / 0.9227& \\
 & H$\rightarrow$B$\rightarrow$R & 30.57 / 0.9244& \\
 & H$\rightarrow$R$\rightarrow$B & 30.34 / 0.9219& \\
 & \bf{R$\rightarrow$B$\rightarrow$H} & 30.79 / 0.9246& \\
 & R$\rightarrow$H$\rightarrow$B & 30.29 / 0.9201& \\

 \hline
 & & &  \\
\end{tabular} &
\begin{tabular}{c|c|c|c}
 & Alignment & PSNR / SSIM & \\
\hline
 & BH$\rightarrow$R & 30.24 / 0.9217& \\
 & B$\rightarrow$HR & 30.36 / 0.9216& \\
 & HR$\rightarrow$B & 30.02 / 0.9189& \\
 & H$\rightarrow$RB & 30.41 / 0.9236& \\
 & RB$\rightarrow$H & 30.47 / 0.9219& \\
 & R$\rightarrow$BH & 30.38 / 0.9228& \\
 & RBH & 30.32 / 0.9206& \\
 \hline
 & & & \\
\end{tabular}
\end{tabular}
}
\end{table}

\begin{table}[t]
\centering
\caption{B: motion-blur removal, H: haze removal, and R: rain-streak removal. The values (PSRN/SSIM are averaged accuracy over the three tasks.}
\label{tab:output_branch4}
\resizebox{\columnwidth}{!}{
\begin{tabular}{cc}
\begin{tabular}{c|c|c|c}
 & Alignment & PSNR / SSIM & \\
 \hline
 & J$\rightarrow$R$\rightarrow$B$\rightarrow$H& 30.09 / 0.8997& \\
 & R$\rightarrow$J$\rightarrow$B$\rightarrow$H& 30.18 / 0.9004& \\
 & \bf R$\rightarrow$B$\rightarrow$J$\rightarrow$H& 30.22 / 0.8995& \\
 & R$\rightarrow$B$\rightarrow$H$\rightarrow$J& 30.17 / 0.9004& \\ 
 \hline
 & & &  \\
\end{tabular} &
\begin{tabular}{c|c|c|c}
 & Alignment & PSNR / SSIM & \\
\hline
 &JR$\rightarrow$B$\rightarrow$H& 30.20 / 0.8985 & \\
 &R$\rightarrow$BJ$\rightarrow$H& 30.13 / 0.8991 & \\
 &R$\rightarrow$B$\rightarrow$HJ& 30.02 / 0.8995 & \\
 \hline
 & & & \\
\end{tabular}
\end{tabular}
}
\end{table}

\subsection{Specific Design for Each Degradation Factor}

Based on the above experiment (i.e., Table \ref{tab:output_branch4}), we use the input/output pair $(I_1,O_1)$ for rain-streak, $(I_2,O_2)$ for motion blur, $(I_3,O_3)$ for JPEG artifacts, and $(I_4,O_4)$ for haze, respectively, as shown in Fig.~\ref{fig:RMBN}. We explain the design of each branch pair below.

\subsubsection{Rain-streak} 
We adopt the design of [51]. 
The input branch $I_1$ consists of a ConvLSTM module and two convolutional layers. The ConvLSTM module consists of a convolutional layer with 32 channels and a LSTM, which has four units with 32 channels. The subsequent two convolutional layers have 64 and 96 channels, respectively, in between which there is a $\times 1/2$ down-sampling layer. We use $3\times3$ kernels for all the convolutional layers in $I_1$. 
%A down-sampling operation is set between the two convolutional layers, we use the same approach for this operation as motion blur removal.
The output branch $O_1$, which consists only of $g_1$, is designed to be symmetric to $I_1$;
%We remove one set of [up-sampling+conv.] from the minimal design (Sec. 3.2 in the main-text) so that 
$g_1$ consists of a [conv.+up-sampling] and two conv. layers. These three conv. layer have 96, 48, and 3 channels, respectively.
% {\color{red} We set channel numbers to be 96, 48 and 3 for the three convolutional layers. [Grammar? Xing: I have updated the red sentence]}

\subsubsection{Motion Blur}
We adopt the design of [17]. The input branch $I_2$  contains three internal branches, so does $O_2$. The whole network is recurrently used (within the motion blur removal task) by using these three internal input-output pairs. Specifically, the input image is first down-sampled by the factor of 1/4 and inputted to one of the three. The output from the corresponding internal output branch is then fed back to the second internal input branch, where it is concatenated in the channel dimension with the original image down-sampled by the factor of 1/2. This is repeated once more, yielding the final result. 

As shown in Fig.~\ref{fig:RMBN}, $I_2$ (after the down-sampling) starts with three parallel conv. layers followed by two conv. layers and a $\times 1/2$ down-sampling layer. These conv. layers have 32 channels. For down-sampling, we use Pixel-unshuffle\footnote{ https://github.com/pytorch/pytorch/issues/2456}. After these, $I_2$ proceeds with another set of three parallel conv. layer, two conv. layer, and a $\times 1/2$ down-sampling layer. These conv. layers all have 64 channels. Finally, $I_2$ ends with another set of three parallel conv. layers with 96 channels. $I_1$ has 13 conv. layers in total, each of which has $3\times 3$ kernels and a subsequent ReLU layer.

% We use 13 convolutional layers for the input branch. Three of them (with 32 channels) are employed in parallel to each other at the beginning of the branch. They are used independently to receive the coarse (=1/4 of image's original size), middle (=1/2 of images' original size) and fine (= image's original size) scales' versions of an input image respectively. Then, two layers (with 32 channels) are employed after them, sharing parameters for all the scales. 
% A down-sampling (Pixel-unshuffle\footnote{ https://github.com/pytorch/pytorch/issues/2456}) is conducted after this five layers. 
% We repeat such design again, with the layers' channel number$=64$, to process the data into a smaller size (1/4 to the input's size). Finally, we end this branch with the rest three layers (with 96 channels) employed independently for the three scales. We use 3$\times$3 kernels for all these layers, and we set a ReLU layer after each convolutional layer. }

The output branch $O_2$ consists of one DuRB-M block and $g_2$, and the latter has a symmetric structure to $I_2$. 
%For updating the output branch, we make it symmetric to the input branch. 
As shown in Fig.~\ref{fig:RMBN}, $g_2$ starts with three parallel paths consisting of a $\times 2$ up-sampling layer with a built-in conv. layer (w/ 96 channels) and a conv. layer (w/ 64 channels) followed by two additional conv. layers (w/ 64 channels). The same structure is repeated once (channels of the conv. layers are 64, 32, and 32, respectively), connecting to the output. 
%We first employ three sets of {\color{red} [up-sampling\footnote{It consists of a convolutional layer with $n$ channels and a pixel-shuffle process. $n=96$ and $64$ for the first and second applications in the output branch}+conv.(with 64 channels)]} for each internal branch. %scales respectively.
%They do not share parameters. These are followed by two convolutional layers (with 64 channels) whose weights are shared between the internal branches. 
%This design repeats again with channel number$=32$ for the convolutional layers, and the output branch ends with a convolutional layer (channel number$=32$) with its parameters shared for the scales. 
All the conv. layers use 3$\times$3 kernels. A ReLU layer follows every conv. layer except the one employed in each up-sampling module.

% In a single forward computation, the coarse scale version (down-scaled by a factor of 4 from the original size) of an training image is firstly fed into a network to have the deblurred result. Then, the result is scaled up by a factor of 2 to the middle scale, and is re-fed into the network with the middle scale version of the training image. This procedure repeats again for the fine scale version to have the final deblurring result.

\subsubsection{JPEG Artifacts} 
The output branch $O_3$ consists of a series of two DuRB-M blocks followed by $g_3$. We use the same design for $I_3$ and $g_3$ as $I_2$ and $g_2$, respectively. We found this works well for this factor. 

\subsubsection{Haze} The output branch $O_4$ consists of a series of three DuRB-M blocks followed by $g_4$.
For $I_4$ and $g_4$, we employ the minimal design explained in the main paper, as shown in Fig.~\ref{fig:RMBN}.

%---------------------------------------------------------------

\begin{table}[t]
\centering
\caption{The number of parameters and the memory footprint of the three sections of the proposed network, i.e., the input branches, the stem network, and the output branches, for the model with the minimal design of the input/output branches and the model with the improved design.} 
\label{tblparam1}
\small
\begin{tabular}{c|c|c|c}
\hline
  \# of param. &input & stem & output  \\
  \hline
  minimal design& 0.07M& 3.94M& 0.52M \\
  \hline
  improved design& 2.35M& 3.94M & 4.96M \\
  
  \hline \hline
  
  GPU memory &input & stem & output  \\
  \hline
  minimal design& 0.3MB & 15.76MB & 2.08MB \\
  \hline
  improved design& 9.43MB& 15.76MB& 19.87MB\\
  \hline
\end{tabular}
\end{table}

\begin{table*}[t]
\centering
\caption{The number of parameters and the memory footprint of each component of the proposed network. M$_i$ indicates a DuRB-M block.}
%{\color{red} [Xing: full version of the table, M$_{i}$ means a DuRB-M.]}} 
\label{tblparam2}
\resizebox{\textwidth}{!}{
\begin{tabular}{cc}
\begin{tabular}{c|c||c|c}

&minimal & \# of param.  & memory \\
\hline
\multirow{4}{*}{in-branch}
         & \multirow{4}{*}{conv.$\times$3}& \multirow{4}{*}{0.07 (M)}&  \multirow{4}{*}{0.3 (MB)}\\
         & & & \\
         & & & \\
         & & & \\
\hline
\multirow{5}{*}{stem}
 &M$_{1}$& 0.64& \multirow{5}{*}{15.76}\\
 &M$_{2}$& 0.74& \\
 &M$_{3}$& 0.74& \\
 &M$_{4}$& 0.84& \\
 &M$_{5}$& 0.98& \\
\hline
\multirow{4}{*}{out-branch} 
& g$_{1}$& 0.13& \multirow{4}{*}{2.08 (=0.52$\times4$)} \\
& g$_{2}$& 0.13& \\
& g$_{3}$& 0.13& \\
& g$_{4}$& 0.13& \\
\hline
\end{tabular} &
\begin{tabular}{c|c||c|c}
  &improved & \# of param. & memory \\
  \hline
\multirow{4}{*}{in-branch}
         & $I_{1}$& 0.32 (M)& {1.26 (MB)}\\
         & $I_{2}$& 0.98& 3.93\\
         & $I_{3}$& 0.98& 3.93\\
         & $I_{4}$& 0.07& 0.3\\
  \hline
 \multirow{5}{*}{stem}
 &M$_{1}$& 0.64& \multirow{5}{*}{15.76}\\
 &M$_{2}$& 0.74& \\
 &M$_{3}$& 0.74& \\
 &M$_{4}$& 0.84& \\
 &M$_{5}$& 0.98& \\
\hline
\multirow{4}{*}{out-branch} 
& g$_{1}$& 0.04& 0.16\\
& M$_{6}+$g$_{2}$& 0.98 + 0.48 & 5.86\\
& M$_{6}+$M$_{7}+$g$_{3}$& 0.98+1.43+0.48& 11.56\\
& M$_{6}+$M$_{7}+$M$_{8}+$g$_{4}$& 0.98+1.43+1.43+0.13& 15.87\\
\hline
\end{tabular}
\end{tabular}
}
\end{table*}

\subsection{Number of parameters}
Our network is composed of the input branches, the stem network, and the output branches. The motivation behind this composition is to make it possible to deal with multiple degradation factors with a compact network. The network is recurrently used, in which each input/output branch pair is specifically used for a single factor while the stem network is used for all the factors. 
%Thus, from the memory efficiency perspective, the stem network should use a certain amount of the memory on a device. 
Table \ref{tblparam1} shows the number of parameters and memory footprint for the three sections for two different designs, i.e., the minimal design and the improved design; the latter was used for the experiments of Sec. 5.3 and 5.4. Table \ref{tblparam2} shows those for each component of the networks. It is seen from Table \ref{tblparam1} that 
the improved design of input/output branches needs a lot more parameters and memory as compared with the minimal design; however they are still comparable to those needed by the stem network. Considering that the network deals with the four different degradation factors, we think this is acceptable.

%the proposed design attains a balance between efficient memory usage and the ability of dealing with multiple degradation factors with a single network.

\section{Details of Training}
\label{sec:train_details}
\paragraph{Global Settings} We use the Adam optimizer with $(\beta_{1}, \beta_{2})=(0.9, 0.999)$ and $\epsilon=1.0 \times 10^{-8}$ for training all the  models. For loss functions, we use a weighted sum of SSIM and $l_{1}$ loss, specifically, $1.1\times$SSIM$+0.75\times l_{1}$ following Liu \etal's setting [42], for all the tasks. The initial learning rate is set to 0.0001 for all the tasks. All the experiments are conducted using PyTorch. Our code and trained models is publicly available (https://github.com/6272code/6272-code). For data augmentation, we use horizontal flip for the removal of motion blur, rain-streak, and JPEG artifacts; for haze, we use horizontal and vertical flips, and image rotations with a degree randomly chosen from $[90, 180, 270]$. 

\paragraph{Training MBN (Sec. 5.3)} 
We set the size of minibatch to $40$ for training the network. We reduced learning rate by $\times0.1$ twice at the iterations where the training loss stopped decreasing. The model is trained $1.2 \times 10^{5}$ iterations. By an iteration, we mean a cycle  (i.e., a set of multiple minibatches containing different factors), as explained in Sec. 4.1. The crop size is $128\times128$ and $256\times256$ pixels for rain-streak and the other factors, respectively.

\paragraph{Training R-MBN (Sec. 5.3)} We fine-tuned the MBN model. The learning rate was set to be 0.000001 for the training. We fixed parameters of all the input branches to accelerate training. The crop size is $128\times128$  and $256\times256$ pixels for rain-streak and the other factors, respectively.
We stopped the fine-tuning when the training loss stopped decreasing. The size of minibatches was set to $4$.

\paragraph{Configuration for Systematic Search of Output Branches}
The models are trained for $1\times 10^{5}$ iterations  with batchsize $= 3$. We resize the images of GoPro dataset and DIV to $\times 1/2$ for training to reduce memory usage. Note that this configuration is used only here. We use the crop size of $256\times256$ pixels for all the models.

\paragraph{Training Details for Ablation Tests1 (Sec. 5.1)}
% In this experiment, we build the input branch and the output branches the same things we used for systematic search (Sec.~\ref{sec:decoder_algin}). We apply the optimal order (\ie R$\rightarrow$B$\rightarrow$J$\rightarrow$H) for aligning the output branches. 
We employ the model having the best design of output branches found by the systematic search.
%explained in {\color{red} Sec.~xx}.
%By disabling/enabling each component of DuRB-M's, we evaluate the contribution of each. 
% The results are shown in Table. 1 in the main-text. 
We train all the models for $5 \times 10^{4}$ iterations with batchsize  $=20$. 
%It is noteworthy that although we have trained these networks less compared to the MBN model used in Table 3 which is trained $1.2 \times 10^{5}$ iterations, we made sure that they have converged as they have less parameters and simpler architectures. We made sure that this difference of training time will not invalidate the conclusion. 
We set crop size to $256\times256$ pixels for all the models here.

\paragraph{Training Details for Ablation Tests2 (Sec. 5.2)}
We use batchsize $= 4$ and learning rate $= 0.000001$. All the models are fine-tuned for $1\times 10^{4}$ iterations starting from the checkpoint at the $8\times 10^{4}$ iteration of training of MBN used in Table 3. For crop size, we use $256\times256$ pixels for motion blur, haze and JPEG artifacts, and $128\times128$ for rain-streak. 

\paragraph{Training Details for Removal of Unknown Combined Degradations}
We have trained two versions of the proposed model, i.e., R-MBN and R-MBN*. For R-MBN, we trained it on uncombined single-factor data with the proposed two-step training method. The crop size is $128\times128$ pixels and $256\times256$ pixels for rain-streak and the other factors, respectively. The model was trained for $8.5\times 10^{4}$ iterations in the first step, and for $2.5\times 10^{4}$ iterations in the second step. For R-MBN*, we fine-tuned the model on the training set of combined factors with crop size of $256\times256$ pixels, starting from a checkpoint of the training of R-MBN. We fixed the parameters of all the input branches to accelerate training. This version was also trained for $2.5\times 10^{4}$ iterations.

\section{Data Used for Experiments on Unknown Combined Degradations (Sec. 5.4)}

As noted in the main paper, we created a dataset for the experiments on removal of unknown combined degradation factors. This dataset is created from the 8,970 outdoor scenes provided in RESIDE-$\beta$ [34]. We split the scenes into 8,478 and 492 and use the former for training and the latter for test. The test set contains the same images used in the synthetic-objective-test (SOTS), which is a popular benchmark test set of RESIDE. We then synthesize pure degradation on the aforementioned training and test images, yielding two training subsets and 16 test subsets, as explained in Sec. 5.4 of the main paper.

\paragraph{Synthesis of  Motion Blur} We employ Hendrycks \etal's approach [24] to synthesize motion blur. The (radius, sigma) used in the approach are randomly chosen from $\{(10,1.5), (10,2.5), (12,3), (12,3.5), (13,8), (15,12)\}$ for each input. The angle of motion blur is randomly generated according to the uniform distribution with the range [-45,45] (in degrees). 
%We use numpy.random.uniform(-45, 45) to decide the motion blur's angle for each input.

\paragraph{Synthesis of Rain-streak} We followed a tutorial\footnote{https://www.youtube.com/watch?v=tYCn8djcI9E} on the creation of rain-streak effects. The rain-streaks are synthesized by motion-blurring Gaussian noise on an image by GIMP on Ubuntu 18.04.2. For each image, we randomly choose a noise level from $[0.55, 0.65, 0.7, 0.8]$ for generating Gaussian noise; we decide the magnitude of motion blur (motion blur length) by randomly selecting a value from $\{10, 20, 35, 50\}$. The direction of rain-streaks for an image is randomly chosen by randomly selecting a motion blur's angle from  $\{110, 105, 100, 90, 85, 80, 70\}$ (in degrees). 

\paragraph{Synthesis of JPEG Artifacts} We simply compress each image by JPEG.  To be specific, we use Pillow\footnote{https://pillow.readthedocs.io/en/stable/index.html}, i.e.,  $\mathtt{PIL.Image.save(buff, quality=X)}$. The compression quality is chosen for each image from the range of $[15, 16, \dots, 50]$.

\paragraph{Synthesis of  Haze} RESIDE-$\beta$ consists of a large number of clean images and their hazy versions with different haze levels. There are two parameters ($v_1$ and $v_2$) used for generating hazy effects in the RESIDE-$\beta$ dataset: $v_1\in \{1, 0.95, 0.9, 0.8, 0.85\}$ and $v_2\in \{0.2, 0.16, 0.12, 0.1, 0.08, 0.06, 0.04\}$. For each clean image, there are 35 ($=5\times7$, all the combinations of $v1$ and $v2$) hazy versions in the dataset. We use these images for our purpose; specifically, for each clean image, we randomly choose $v_1$ from the above set, and $v_2$ from $\{0.2, 0.16, 0.12, 0.1, 0.08\}$.
%, and employ the corresponding hazy image of RESIDE-$\beta$ for our dataset. 

\begin{figure*}[bt]
\centering
\includegraphics[width=.85\textwidth]{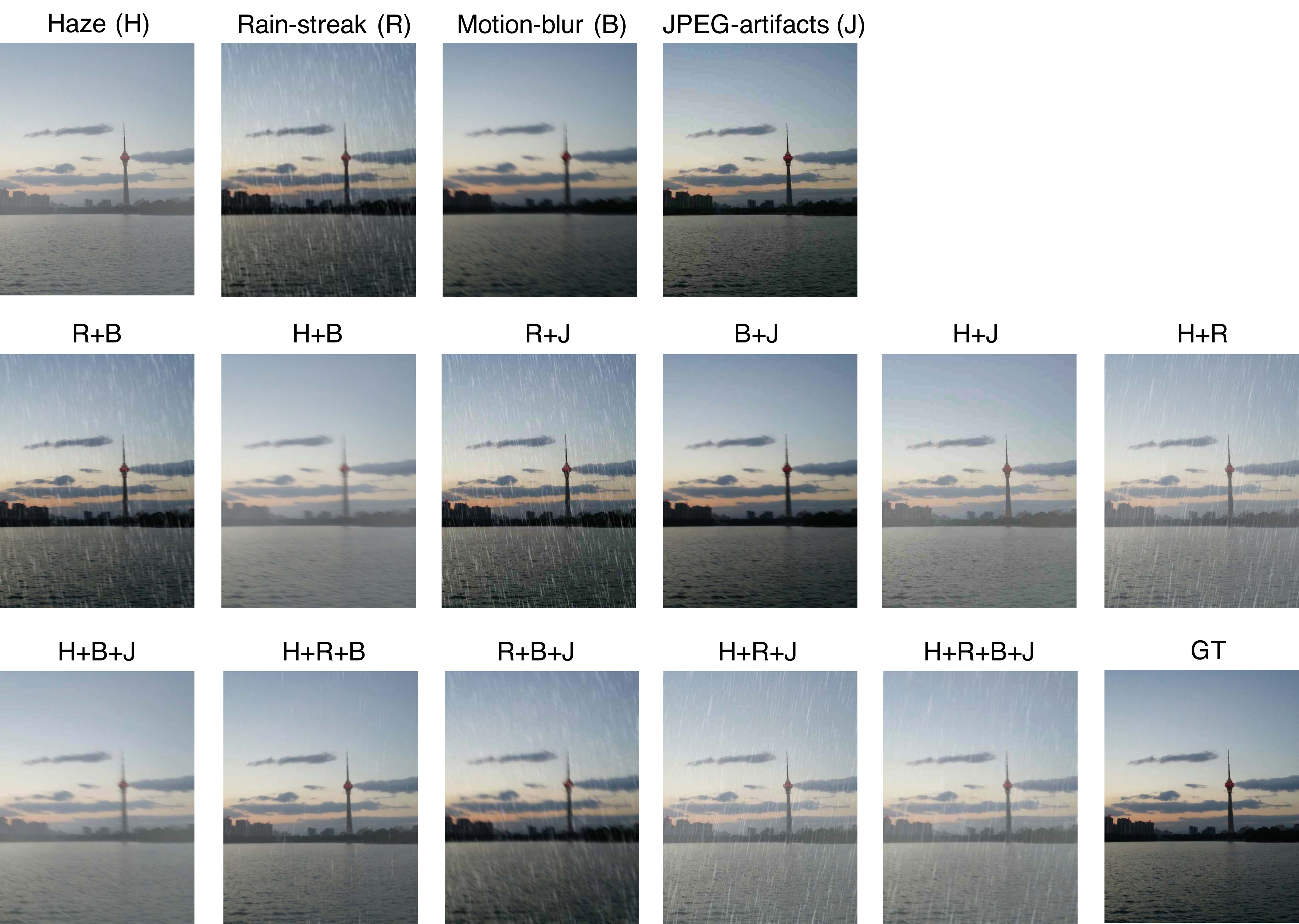}
\caption{Examples of the images with synthetic degradation. These are used in the experiment of Sec.~5.4 of the main paper. }
\label{fig:mymix1}
\end{figure*}

\begin{figure*}[bt]
\centering
\includegraphics[width=\textwidth]{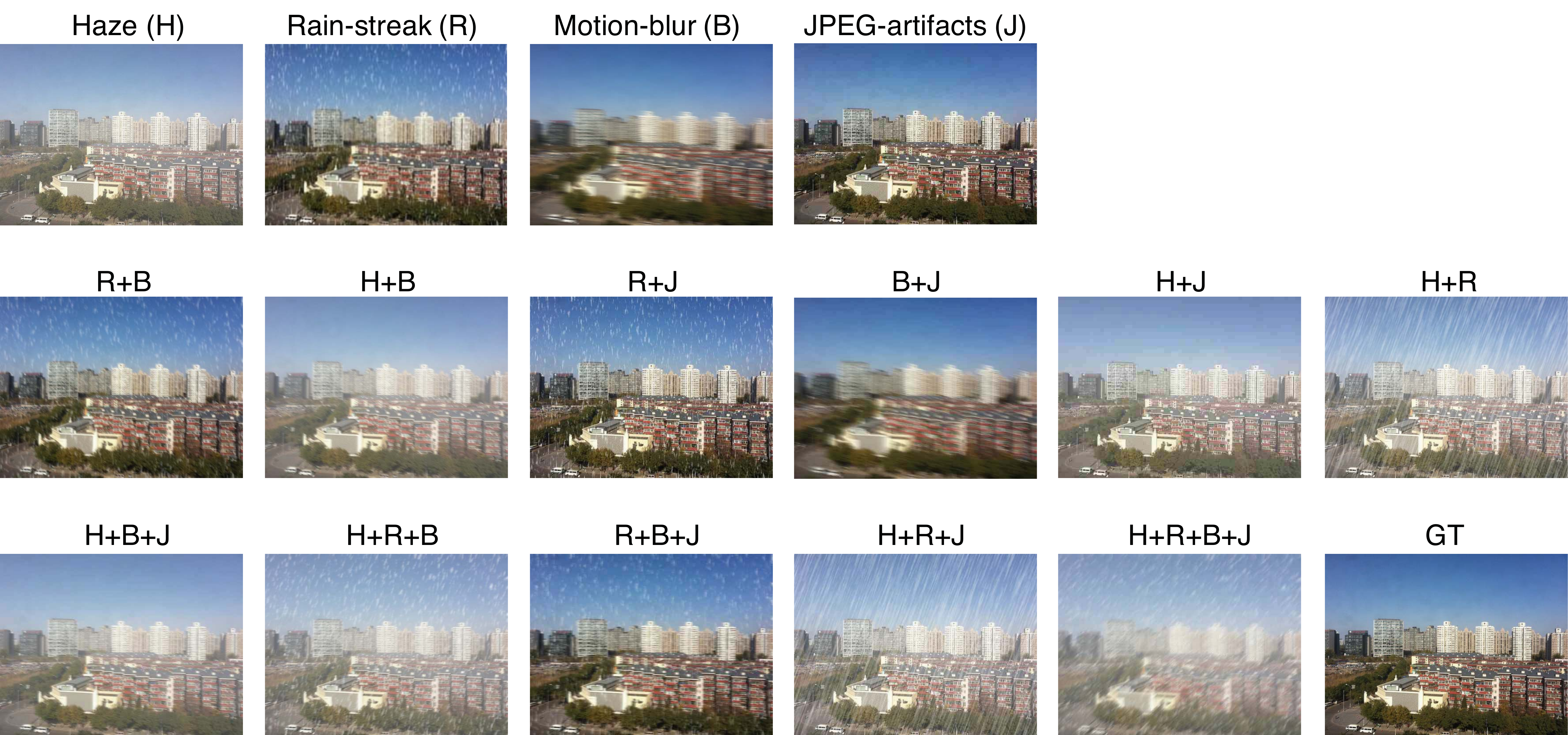}
\caption{Examples of the images with synthetic degradation. These are used in the experiment of Sec.~5.4 of the main paper. }
\label{fig:mymix2}
\end{figure*}

\paragraph{Order of Synthesis} 
When applying the above effects on an image, their order is important. Considering a natural process of image formation, we employ the following order: haze, rain-streak, motion blur, JPEG artifacts. This order applies to all the cases of combining two, three, four, and random number factors. Examples of the synthesized images are shown in Fig.~\ref{fig:mymix1} and \ref{fig:mymix2}. 

%\section{Examples of Output Images}

\section{Examples of Restoration from Combined Degradation (Sec.~5.4)}

\begin{figure*}[bt]
\centering
\includegraphics[width=\textwidth]{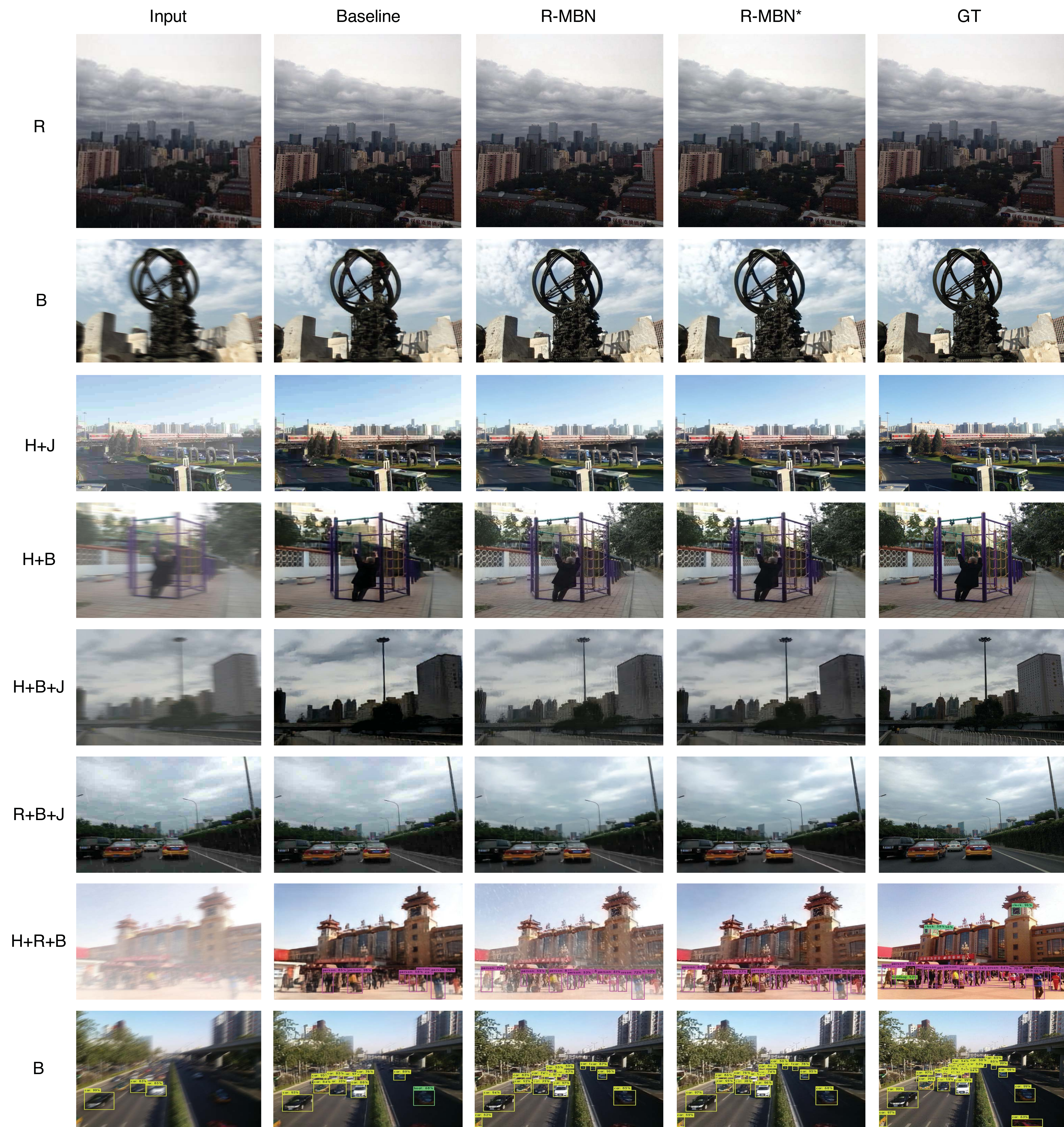}
\caption{Qualitative evaluation of the proposed method (R-MBN and R-MBN*) and the compared baseline (i.e., Suganuma \etal [58]). R, B, J and H means rain-streak, motion blur, JPEG artifacts, and haze. R-MBN and R-MBN* are the proposed models. Object detection is performed by YOLO-v3.}
\label{fig:mymix3}
\end{figure*}

In Sec.~5.4 of the main paper, we evaluate the performances of several methods using the dataset created as above. Figure \ref{fig:mymix3} shows several examples of the results obtained by the proposed methods (R-MBN and R-MBN*) and the baseline (Suganuma et al. [58]), which are omitted in the main paper due to lack of space. R-MBN is trained on images with a single degradation factor, whereas R-MBN* is trained on images with mixed degradation factors. In the last two rows of the figure, we show the outputs of an object detector (YOLO-v3\footnote{https://pjreddie.com/darknet/yolo/}) applied to each image, i.e., the original input, the restored images by the methods, and the ground-truth image. 

\end{document}